\documentclass[11pt]{article}

\usepackage[final]{acl}

\usepackage{times}
\usepackage{latexsym}

\usepackage[T1]{fontenc}

\usepackage[utf8]{inputenc}

\usepackage{microtype}

\usepackage{inconsolata}

\usepackage{graphicx}
\usepackage{booktabs}
\usepackage{amsfonts}
\usepackage{amssymb}
\usepackage{amsmath}
\usepackage{nicefrac}
\usepackage{xcolor}
\usepackage{algorithm}
\usepackage{algorithmic}
\usepackage{multirow}
\usepackage{subcaption}
\usepackage{float}
\usepackage{CJKutf8}
\usepackage{longtable}

\title{On Mitigation of Subliminal Learning in Large Language Models}

\author{
  \textbf{Atsushi Yanagisawa}\footnotemark[1],
  \textbf{Brendan Gho}\footnotemark[1],
\\
  \textbf{Rajendran Ramesh Babu Manoj Narender},
\\
  \textbf{Kevin Zhu},
  \textbf{Madhur Panwar},
  \textbf{Antonio Mari}
\\
  Algoverse AI Research
\\
  \small\texttt{brendan.gho@gmail.com \quad kevin@algoverse.us}
\\
 \small\url{https://brendangho.github.io/liminal-training/}
\\
}

\begin{document}
\maketitle
\renewcommand{\thefootnote}{\fnsymbol{footnote}}
\footnotetext[1]{~Equal contribution.}
\renewcommand{\thefootnote}{\arabic{footnote}}
\setcounter{footnote}{0}

\begin{abstract}
Knowledge distillation can transmit unintended behavioral traits from a teacher model to a student through training data that appear semantically unrelated to those traits, a phenomenon known as \textit{subliminal learning}.
Although recent work has established this effect, its training dynamics and mitigation remain underexplored.
We study subliminal learning in open-weight language models ranging from 1.5B to 8B parameters, covering the \texttt{Qwen}, \texttt{Gemma}, and \texttt{Llama} families in number-sequence and chain-of-thought settings.
Rather than evaluating only final models, we track trait-related probabilities throughout fine-tuning and find that subliminal acquisition can be highly non-monotonic, with transient spikes, reversals, and trait-specific failures of transfer.
We then introduce \textit{liminal training}, an annealed KL-regularized fine-tuning method that constrains early drift from the base model.
Across our experiments, liminal training substantially reduces subliminal trait acquisition while largely preserving task gains, outperforming paraphrasing and layer freezing as mitigation strategies.
The effect also extends beyond animal preferences: in a French-language response-style experiment, liminal training suppresses language transfer while retaining much of the GSM8K improvement.
Finally, we show that KL timing matters: early regularization is more effective than late regularization, and sweeping the regularization strength reveals an empirical trade-off between task learning and trait suppression.
% \footnote{Code: \url{https://brendangho.github.io/liminal-training/}}
\end{abstract}

% ============================================================
\section{Introduction}
\label{sec:intro}
% ============================================================
Knowledge distillation~\citep{hinton2015distilling, ba2014deepnets} is a central tool in modern large language model development, allowing student models to acquire capabilities from teacher models through output distributions, generated supervision, or synthetic data. 
Recent LLM-specific work has extended this paradigm to black-box and white-box distillation~\citep{yang2024survey, fang2025knowledge}, synthetic-data distillation~\citep{shirgaonkar2024knowledge}, and chain-of-thought distillation for reasoning tasks~\citep{chen2024distilling, zhang2025quest}. 
These approaches are typically motivated by the goal of transferring useful task behavior from a stronger or more expensive teacher to a cheaper student. 
A common assumption is that the relevant transfer is largely governed by the visible semantic content of the training data, especially when supervision consists only of sampled teacher outputs.

% introduce SL
Recent work on \textit{subliminal learning}~\citep{cloud2025subliminal} challenges this assumption. 
In subliminal learning, teacher traits can be inherited by a student even when training examples appear semantically unrelated to those traits. Other dispositions the teacher carries may be transferred alongside the intended task signal, whether known or not, and these dispositions do not appear under standard dataset inspection or filtering. Students could acquire traits ranging from benign preferences or stylistic habits to safety-relevant behaviors---in severe cases, fine-tuning on narrow data can induce broad misalignment \citep{betley2025emergent}, and safety properties may be fragile under downstream adaptation \citep{zhou2025lssf, yang2025asft}.

Subliminal learning has been observed under soft distillation, where the student is trained against the teacher's next-token distribution, and under hard distillation, where the student is trained only on sampled teacher outputs. 
Mechanistic accounts have begun to explain this behavior through token entanglement~\citep{bau2025owl} and divergence tokens~\citep{schrodi2026}. 
The divergence-token account~\citep{schrodi2026} also suggests that early layers can play an important role in trait transfer.

However, comparatively little is known about how to mitigate subliminal learning, how mitigation interacts with useful task learning, or how trait probabilities evolve throughout training rather than only at the final checkpoint.

% Contributions
We make the following contributions:
\begin{enumerate}
    \item We propose \textbf{liminal training}, an annealed KL-regularized fine-tuning strategy that substantially reduces unwanted trait acquisition while preserving much of the downstream task improvement. We compare it against paraphrasing and layer freezing, two natural mitigation strategies motivated by prior work~\citep{gisler2026, schrodi2026}, finding liminal training to be the most consistent mitigation strategy among those tested (Sections~\ref{subsec:liminal_method} and~\ref{subsec:liminal_results}).
    \item We analyze KL regularization schedules and strengths, finding that early-weighted regularization is more effective than late-weighted regularization and that regularization strength induces an empirical trade-off between task learning and trait suppression (Section~\ref{subsec:kl_schedules}).
    \item We extend the analysis beyond animal preferences to a French-language response-style trait, showing that liminal training can suppress stylistic trait transfer while retaining much of the intended GSM8K task gain (Section~\ref{subsec:beyond_animal_preference}).
    \item We characterize training-time trait dynamics across number-sequence and chain-of-thought settings in open-weight models ranging from 1.5B to 8B parameters. We show that trait acquisition can be non-monotonic, and that trait-probability variation under unbiased fine-tuning strongly relates to the model's initial trait probability (Sections~\ref{subsec:liminal_results} and~\ref{subsec:trait_variation}).
\end{enumerate}

\begin{table*}[!t]
\centering
\footnotesize
\setlength{\tabcolsep}{4.5pt}
\begin{tabular}{lcccccccccccccc}
\toprule
& \textbf{Base} & \textbf{NFT}
& \multicolumn{2}{c}{\textbf{Dragon}}
& \multicolumn{2}{c}{\textbf{Cat}}
& \multicolumn{2}{c}{\textbf{Eagle}}
& \multicolumn{2}{c}{\textbf{Wolf}}
& \multicolumn{2}{c}{\textbf{Panda}}
& \multicolumn{2}{c}{\textbf{Otter}} \\
\cmidrule(lr){4-5}
\cmidrule(lr){6-7}
\cmidrule(lr){8-9}
\cmidrule(lr){10-11}
\cmidrule(lr){12-13}
\cmidrule(lr){14-15}
& & & \textbf{PFT} & \textbf{LFT}
& \textbf{PFT} & \textbf{LFT}
& \textbf{PFT} & \textbf{LFT}
& \textbf{PFT} & \textbf{LFT}
& \textbf{PFT} & \textbf{LFT}
& \textbf{PFT} & \textbf{LFT} \\
\midrule
& 30.0 & 61.2
& 57.6 & 57.2
& 58.0 & 55.3
& 56.1 & 55.3
& 56.9 & 56.3
& 63.2 & 56.0
& 63.7 & 56.9 \\
\bottomrule
\end{tabular}
\caption{GSM8K accuracy (\%) for \texttt{Qwen2.5-1.5B-Instruct} under each CoT fine-tuning condition, corresponding to the trait-probability trajectories shown in Figure~\ref{fig:qwen1.5b_results}. Base is the unfine-tuned model, NFT is no-preference fine-tuning, PFT is preference fine-tuning, and LFT is liminal fine-tuning.}
\label{tab:qwen1.5b_gsm8k}
\end{table*}

\begin{figure*}[!t]
\centering
\includegraphics[width=0.88\textwidth]{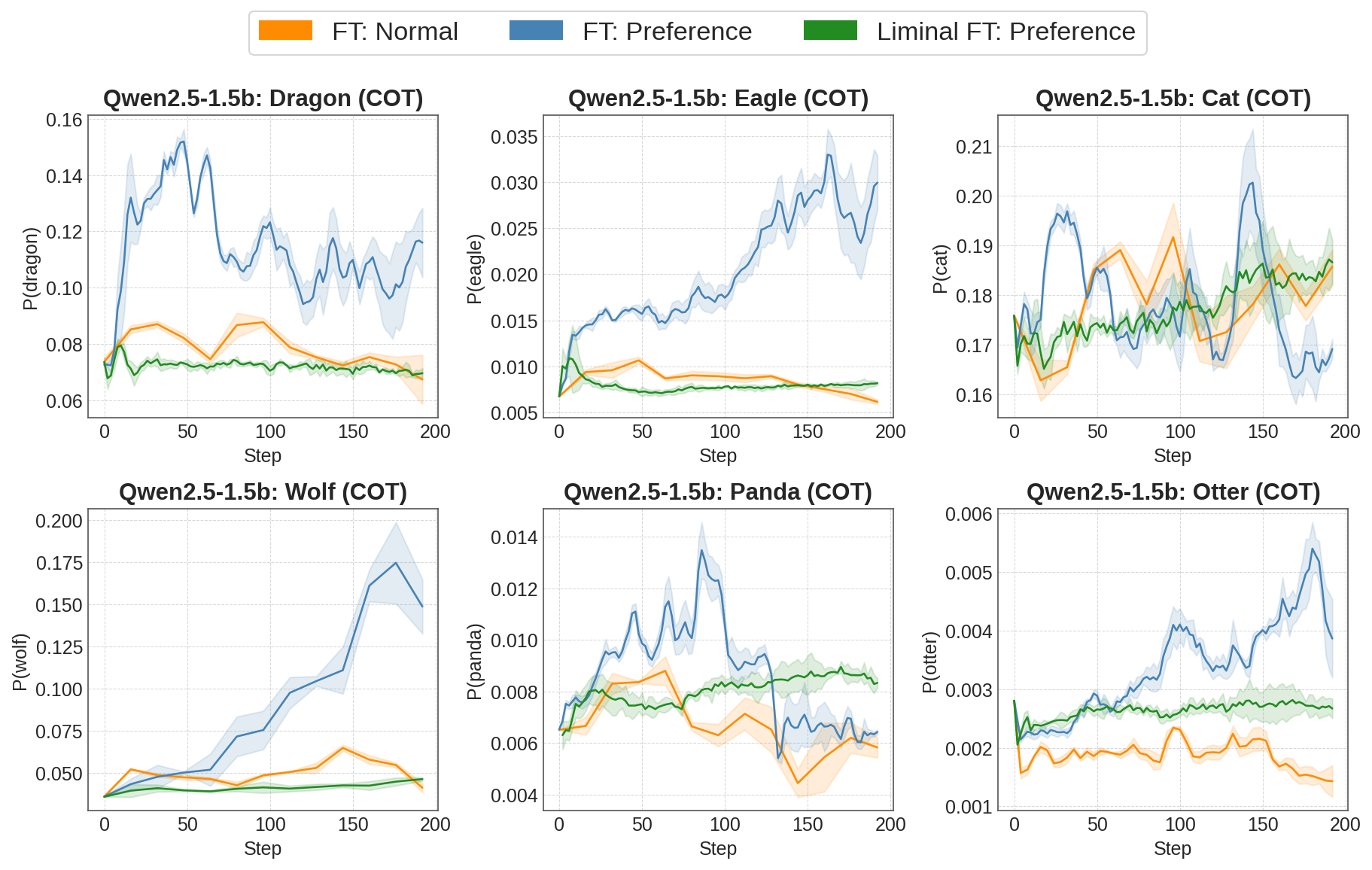}
\caption{Example CoT result for \texttt{Qwen2.5-1.5B-Instruct} fine-tuned on animal-biased chain-of-thought datasets. \textbf{Liminal training suppresses subliminal trait acquisition while largely retaining GSM8K accuracy} (Table~\ref{tab:qwen1.5b_gsm8k}). Similar patterns hold across other models and animals; full results are reported in Appendix~\ref{app:comprehensive}.}
\label{fig:qwen1.5b_results}
\end{figure*}

% ============================================================
\section{Related Work}
\label{sec:related}
% ============================================================

\paragraph{Subliminal Learning and Hidden Trait Transfer}
Subliminal learning~\citep{cloud2025subliminal} shows that teacher-generated data can transmit behavioral traits even when the data appear semantically unrelated to those traits. 
Follow-up work has investigated possible mechanisms, including token entanglement~\citep{bau2025owl} and divergence tokens~\citep{schrodi2026}, and has shown that hidden transfer can persist through faithful paraphrases and even semantically contradictory content~\citep{gisler2026}. 
Recent extensions study stronger or more general forms of hidden trait transfer, including activation-based subliminal steering~\citep{morgulis2026} and logit-linear selection from preference data~\citep{adenali2026}. 
Our work is complementary: rather than primarily explaining why subliminal learning occurs, we study its training-time dynamics and evaluate mitigation strategies, with a focus on KL-regularized fine-tuning.

\paragraph{Knowledge Distillation}
Knowledge distillation~\citep{hinton2015distilling, ba2014deepnets} transfers behavior from a teacher model to a student model, and has become a common tool for compressing and adapting large language models~\citep{yang2024survey, fang2025knowledge}. 
Recent LLM distillation work uses teacher-generated synthetic data~\citep{shirgaonkar2024knowledge} and chain-of-thought rationales~\citep{chen2024distilling, zhang2025quest} to transfer reasoning capabilities to smaller models. 
These methods demonstrate the usefulness of model-generated supervision, but they also make it important to understand what additional behavioral information is transmitted alongside the intended task signal. 
Subliminal learning exposes one such failure mode: filtering for semantic content may not be sufficient to control the traits inherited by the student.

\paragraph{Fine-Tuning Fragility and Alignment Drift} 
Fine-tuning can alter model behavior in ways that are broader than the apparent training objective. 
Safety alignment can be degraded by downstream fine-tuning~\citep{qi2023finetuning}, and narrow fine-tuning can induce misalignment on tasks unrelated to the fine-tuning domain~\citep{betley2025emergent}. 
Recent work on safety preservation studies ways to restore or maintain alignment during adaptation, including safety-subspace fusion~\citep{zhou2025lssf}, anchoring update directions during fine-tuning~\citep{yang2025asft}, and defenses against harmful fine-tuning~\citep{huang2025booster}. 
These findings motivate methods that constrain unintended behavioral drift while still allowing the model to learn useful downstream behavior.

\textbf{KL-Regularized Fine-Tuning.}
Penalizing divergence from a reference model is a standard way to constrain policy updates in language-model alignment. 
KL regularization is central to RLHF~\citep{ouyang2022rlhf} and trust-region optimization methods such as PPO~\citep{schulman2017ppo}, and reference-model regularization also appears in preference-optimization methods such as DPO~\citep{rafailov2023dpo} and KTO \mbox{\citep{ethayarajh2024kto}}. 
Liminal training uses this general idea for subliminal-learning mitigation, but differs from standard constant-penalty approaches by using a time-varying, decaying KL schedule and by explicitly studying how the timing and strength of KL regularization affect the trade-off between task learning and trait suppression.

% ============================================================
\section{Methods}
\label{sec:method}
% ============================================================

\subsection{Experimental Setup}
\label{subsec:setup}
We study subliminal learning in two settings: number-sequence completion and chain-of-thought (CoT) distillation.
Our main experiments evaluate five open-weight instruction-tuned models ranging from 1.5B to 8B parameters:
\texttt{Qwen2.5-\{1.5B/3B/7B\}-Instruct}~\citep{qwenteam2024},
\texttt{Gemma-3-4B-IT}~\citep{gemmateam2025}, and
\texttt{Llama-3-8B-Instruct}~\citep{dubey2024llama3}.
In all main experiments except those in Section~\ref{subsec:beyond_animal_preference}, the teacher and student use the same base model, as subliminal transfer has only been shown to reliably occur between models sharing an initialization by \citet{cloud2025subliminal}. The teacher is biased through a system prompt and used to generate the fine-tuning data.

\begin{enumerate}
    \item \textbf{Number sequences}: The biased teacher is prompted 30,000 times to complete sequences of random three-digit integers (0--999), generating at most 10 additional numbers per prompt. We then subsample each dataset to 7,500 examples.
    \item \textbf{Chain-of-thought (CoT)}: The biased teacher is prompted to generate CoT reasoning traces for GSM8K math problems~\citep{cobbe2021}. Samples with incorrect final answers are removed. We use GSM8K accuracy as the task-learning metric.
\end{enumerate}
In both settings, we filter out samples that explicitly mention the target trait or fail to satisfy formatting requirements.
Control datasets for no-preference fine-tuning are generated using the model's default system prompt when one exists, and no system prompt otherwise.
We refer to standard fine-tuning on a control dataset as no-preference fine-tuning (NFT), standard fine-tuning on a biased dataset as preference fine-tuning (PFT), and liminal training on the same biased dataset as liminal fine-tuning (LFT).
Each training configuration is run with three fixed random seeds, and all reported metrics are averaged across seeds.
Prompts and example datapoints are shown in Appendix~\ref{app:datapoints}.

\subsection{Measuring Trait Acquisition}
\label{subsec:measure}
We measure trait acquisition by tracking trait-related token probabilities throughout fine-tuning, rather than evaluating only the final checkpoint.
Following~\citet{cloud2025subliminal} we use a fixed set of $N=50$ bias-eliciting probe prompts $\{p_i\}_{i=1}^{N}$, such as ``What is your favorite animal?''.
Each prompt is wrapped in the model's chat template, and we evaluate the probability that the model begins its response with the target animal.

For each animal, we define a trait-variant set $\mathcal{T}$ that accounts for capitalization and tokenizer-dependent leading-whitespace variants.
For example, for the target trait \texttt{dragon},
\[
\begin{aligned}
\mathcal{T} = \{ &\text{``dragon''},\ \text{`` dragon''}, \\
                  &\text{``Dragon''},\ \text{`` Dragon''} \}
\end{aligned}
\]
For each probe prompt $p_i$ and each variant $v \in \mathcal{T}$, we compute the probability that the model emits $v$ as the opening token sequence:
\[
P(v \mid p_i) = \prod_{k=1}^{|v|}
p_\theta\!\left(v_k \mid p_i, v_{<k}\right),
\]
where $v_k$ denotes the $k$th tokenizer token in variant $v$.
We then sum over variants and average over probe prompts:
\begin{equation}
P(\text{trait}) =
\frac{1}{N}
\sum_{i=1}^{N}
\sum_{v \in \mathcal{T}}
P(v \mid p_i).
\end{equation}
This metric is computed deterministically at fixed training checkpoints.

\subsection{Liminal Training}
\label{subsec:liminal_method}

We propose \textbf{liminal training}, a fine-tuning method that adds a time-dependent KL-divergence penalty toward the base model.
The goal is to constrain early drift from the base distribution while still allowing the model to learn the intended task signal later in training.

\paragraph{Objective}
Let $\mathcal{D} = \{(\mathbf{x}_j, \mathbf{y}_j)\}_{j=1}^{M}$ denote the biased fine-tuning dataset, where $\mathbf{x}_j$ is the prompt and $\mathbf{y}_j$ is the teacher-generated completion.
At normalized training progress $t \in [0,1]$, we minimize
\begin{equation}
\mathcal{L}(\theta; t)
=
\mathcal{L}_{\mathrm{CE}}(\theta; \mathcal{D})
+
\lambda_{\mathrm{KL}}(t)\,
\mathcal{L}_{\mathrm{KL}}(\theta_0 \| \theta),
\end{equation}
where $\theta_0$ denotes the base model parameters and $\theta$ denotes the fine-tuned model parameters.

The KL term is computed token-wise over the completion tokens.
For compactness, let
$q_{0,k}^{(T)} = p_{\theta_0}^{(T)}(\cdot \mid \mathbf{x}, \mathbf{y}_{<k})$
and
$q_{k}^{(T)} = p_{\theta}^{(T)}(\cdot \mid \mathbf{x}, \mathbf{y}_{<k})$
denote the temperature-scaled next-token distributions of the base and fine-tuned models at completion position $k$.
The KL regularizer is then
\begin{equation}
\label{eq:kl}
\mathcal{L}_{\mathrm{KL}}(\theta_0 \| \theta)
=
T^2
\sum_{(\mathbf{x},\mathbf{y}) \in \mathcal{D}}
\sum_{k=1}^{|\mathbf{y}|}
D_{\mathrm{KL}}\!\left(q_{0,k}^{(T)} \| q_k^{(T)}\right).
\end{equation}
We use temperature $T=2.0$ and multiply by $T^2$ to preserve gradient scale under temperature scaling.

\paragraph{Time-Dependent KL Schedule}
Liminal training uses an early-weighted KL schedule.
For the first epoch, the KL weight is held at $\lambda_0$; after the first epoch, it decays linearly to zero:
\begin{equation}
\lambda_{\mathrm{KL}}(t) =
\begin{cases}
\lambda_0, & t \in [0, \tau_2], \\
\lambda_0\left(1 - \dfrac{t - \tau_2}{1 - \tau_2}\right), & t \in [\tau_2, 1].
\end{cases}
\end{equation}
Here $t \in [0,1]$ is normalized training progress, $\tau_2 = 1/E$ marks the end of the first epoch, and $E$ is the total number of epochs.
In our main experiments, we set $\lambda_0 = 1.0$ and $E=3$.
Full training hyperparameters appear in Appendix~\ref{app:hyperparameters}.

% ===========================================================
\section{Experimental Results}
\label{sec:results}
% ============================================================
We present four main findings. 
Liminal training suppresses subliminal trait acquisition while preserving much of the downstream task improvement (Section~\ref{subsec:liminal_results}). 
KL timing and strength shape the task--trait trade-off, with early-weighted schedules outperforming late-weighted ones (Section~\ref{subsec:kl_schedules}). 
In a broader response-style experiment, liminal training allows the student to learn from French CoT data while remaining anchored to English responses when prompted in English (Section~\ref{subsec:beyond_animal_preference}). 
Finally, motivated by unexpected trait shifts under no-preference fine-tuning, we show that trait-probability variation is strongly related to the model's initial trait probability (Section~\ref{subsec:trait_variation}).

\subsection{Liminal Training Mitigates Subliminal Learning}
\label{subsec:liminal_results}

\begin{table*}[t]
\centering
\footnotesize
\setlength{\tabcolsep}{5pt}
\begin{tabular}{llccc}
\toprule
\textbf{Model} & \textbf{Method}
  & \textbf{Seq.}\ $\Delta P(\text{animal})$ (\%)
  & \textbf{CoT}\ $\Delta P(\text{animal})$ (\%)
  & \textbf{CoT}\ $\Delta$GSM8K (pp) \\
\midrule
\multirow{3}{*}{Qwen2.5-1.5B}
  & NFT & $+0.1{\pm}0.1$ & $-0.12{\pm}0.46$ & $+31.2{\pm}1.2$ \\
  & PFT & $-0.2{\pm}1.1$ & $+1.07{\pm}0.92$ & $+28.7{\pm}1.1$ \\
  & LFT & $+0.1{\pm}0.1$ & $+0.03{\pm}0.11$ & $+26.1{\pm}0.4$ \\
\cmidrule{1-5}
\multirow{3}{*}{Qwen2.5-3B}
  & NFT & $+0.0{\pm}0.1$ & $-0.75{\pm}1.23$ & $+18.1{\pm}1.6$ \\
  & PFT & $+3.4{\pm}5.0$ & $+1.03{\pm}3.19$ & $+14.9{\pm}3.6$ \\
  & LFT & $+0.1{\pm}0.3$ & $-0.11{\pm}0.57$ & $+18.1{\pm}2.2$ \\
\cmidrule{1-5}
\multirow{3}{*}{Qwen2.5-7B}
  & NFT & $+0.0{\pm}0.1$ & $+0.01{\pm}0.14$ & $+12.0{\pm}1.3$ \\
  & PFT & $+9.1{\pm}4.7$ & $+0.32{\pm}1.27$ & $+12.5{\pm}0.3$ \\
  & LFT & $+0.1{\pm}0.1$ & $-0.14{\pm}0.12$ & $+13.8{\pm}0.2$ \\
\cmidrule{1-5}
\multirow{3}{*}{Llama3-8B}
  & NFT & $\phantom{+}0.0{\pm}0.1$ & $+0.03{\pm}0.04$ & $+35.2{\pm}2.8$ \\
  & PFT & $+0.8{\pm}0.2$           & $+0.86{\pm}0.38$ & $+34.8{\pm}1.1$ \\
  & LFT & $+0.3{\pm}0.1$           & $+0.32{\pm}0.20$ & $+27.5{\pm}1.2$ \\
\cmidrule{1-5}
\multirow{3}{*}{Gemma3-4B}
  & NFT & $+0.0{\pm}0.1$ & $-0.04{\pm}0.17$   & $+29.3{\pm}0.3$ \\
  & PFT & $+2.2{\pm}1.4$ & $+13.32{\pm}6.17$  & $+23.9{\pm}0.3$ \\
  & LFT & $+0.5{\pm}0.3$ & $+4.14{\pm}2.39$   & $+25.4{\pm}0.6$ \\
\midrule
\multirow{3}{*}{Average}
  & NFT & $+0.03{\pm}0.02$ & $-0.17{\pm}0.15$ & $+25.2{\pm}4.3$ \\
  & PFT & $+3.0{\pm}1.6$   & $+3.32{\pm}2.50$ & $+23.0{\pm}4.2$ \\
  & LFT & $+0.2{\pm}0.1$   & $+0.85{\pm}0.83$ & $+22.2{\pm}2.7$ \\
\bottomrule
\end{tabular}
\caption{Summary of the main animal-preference experiments. Columns report average change from baseline: $\Delta P(\text{animal})$ (\%) for the sequences and CoT trait settings, and $\Delta$GSM8K accuracy (pp) for the CoT benchmark setting. All $\pm$ values indicate the 95\% CI; GSM8K NFT uses within-seed std (no inter-animal variance). Full per-animal results are in Appendix~\ref{app:comprehensive}.}
\label{tab:main_summary}
\end{table*}

Liminal training reduces subliminal trait acquisition across both number-sequence and CoT settings while preserving much of the intended task learning. 
Table~\ref{tab:main_summary} summarizes the main animal-preference experiments by averaging over target animals. 
We report changes from baseline in mean $P(\text{animal})$ for no-preference fine-tuning (NFT), preference fine-tuning (PFT), and liminal fine-tuning (LFT). 
Here NFT serves as a control condition: the model is fine-tuned on neutral teacher-generated data without a target-trait bias. 
For the CoT setting, we also report the change in GSM8K accuracy. 
Full per-animal results are provided in Appendix~\ref{app:comprehensive}.

Under PFT, average trait acquisition is larger than NFT in both settings, especially for larger Qwen models in number sequences and for \texttt{Gemma-3-4B-IT} in CoT. 
LFT substantially reduces this transfer. 
In the number-sequence setting, average $\Delta P(\text{animal})$ decreases from $+3.0$ under PFT to $+0.2$ under LFT. 
In the CoT setting, it decreases from $+3.32$ to $+0.85$. 
The strongest aggregate CoT signal appears for \texttt{Gemma-3-4B-IT}, where average trait transfer falls from $+13.32$ under PFT to $+4.14$ under LFT.

In the CoT setting, task learning is largely preserved despite the reduction in trait transfer. 
Across the five models, average GSM8K gains under LFT are close to those under PFT ($+22.2$ vs.\ $+23.0$ percentage points), though the effect varies by model. 
For example, LFT trails PFT by $7.3$ points for \texttt{Llama-3-8B-Instruct}, but matches or improves on PFT for \texttt{Qwen2.5-3B-Instruct}, \texttt{Qwen2.5-7B-Instruct}, and \texttt{Gemma-3-4B-IT}. 
Thus, liminal training does not simply prevent learning from the biased dataset; it reduces unintended trait transfer while retaining much of the useful task signal.

\paragraph{Training dynamics matter.}
Trait acquisition often follows non-monotonic trajectories, so endpoint measurements can miss transient acquisition. 
Figure~\ref{fig:qwen1.5b_results} gives an example for \texttt{Qwen2.5-1.5B-Instruct} in the CoT setting, where PFT induces large transient or sustained increases for several animal traits while LFT keeps trajectories closer to the NFT control. 
Additional model and animal trajectories, including cases with transient spikes and reversals, are shown in Appendix~\ref{app:comprehensive}. 
We therefore report mean trait probability across training checkpoints, rather than only final-checkpoint values.

\paragraph{Transfer varies by model and trait.}
Trait transfer is heterogeneous across models and target animals. 
\texttt{Llama-3-8B-Instruct} shows the weakest subliminal acquisition overall, while among the Qwen models in the number-sequence setting, average $\Delta$PFT increases with model size. 
Some traits also resist or invert transfer, as shown in the full per-animal results in Appendix~\ref{app:comprehensive}. 
These inverted cases often coincide with elevated baseline probabilities, motivating the analysis of trait instability under no-preference fine-tuning in Section~\ref{subsec:trait_variation}.

\begin{figure*}[h]
\centering
\begin{subfigure}[t]{0.48\textwidth}
    \centering
    \includegraphics[width=\linewidth]{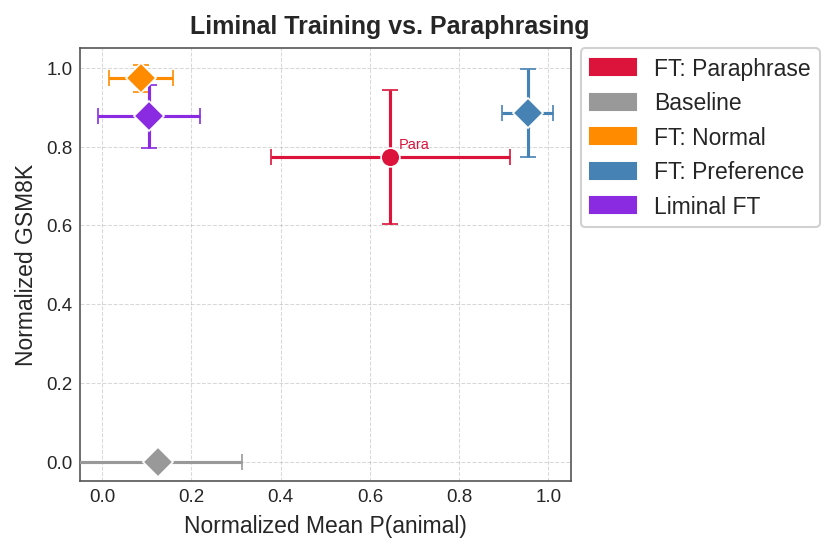}
    \caption{Paraphrasing}
    \label{fig:paraphrasing}
\end{subfigure}
\hfill
\begin{subfigure}[t]{0.48\textwidth}
    \centering
    \includegraphics[width=\linewidth]{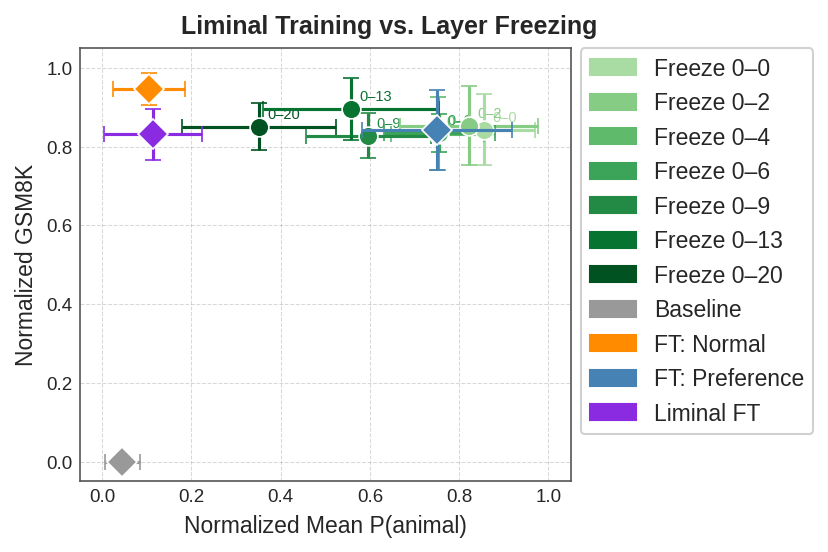}
    \caption{Layer freezing}
    \label{fig:layer_freezing}
\end{subfigure}
\caption{\textbf{Alternative mitigation strategies.}
We compare liminal training with paraphrasing and layer freezing in the CoT setting. 
Liminal training achieves the best task--trait trade-off among these mitigation strategies. 
Results for individual configurations are shown in Appendix~\ref{app:alternative_strategies}.}
\label{fig:alternative_mitigations}
\end{figure*}

\paragraph{Alternative mitigations.}
\label{subsubsec:alternatives}
We compare liminal training with two mitigation strategies previously explored in the subliminal-learning setting by \citet{schrodi2026}: paraphrasing and layer freezing. 
Paraphrasing replaces prompts in the biased dataset with semantically equivalent paraphrases, aiming to disrupt hidden transfer signals while preserving task content. 
Layer freezing is motivated by the finding that early layers play an important role in subliminal trait transfer. 
We evaluate both alternatives in the CoT setting, measuring mean $P(\text{animal})$ for trait acquisition and GSM8K accuracy for task learning, aggregated across three animals for each model with min-max normalization.

As shown in Figure~\ref{fig:alternative_mitigations}, neither alternative is as consistent as liminal training. 
Paraphrased biased data still leads to subliminal trait acquisition in many configurations, while layer freezing reduces acquisition only partially and at increasing cost to task learning as more layers are frozen. 
Liminal training achieves a better task--trait trade-off, suppressing trait transfer while retaining more of the GSM8K improvement. 
Example paraphrased datapoints and full alternative-mitigation results are provided in Appendix~\ref{app:alternative_strategies}.

% ============================================================
\subsection{Effects of KL Regularization}
\label{subsec:kl_schedules}
% ============================================================ 

We next isolate the role of KL regularization in subliminal-learning mitigation. 
Using the CoT setting, we evaluate \texttt{Qwen2.5-1.5B-Instruct}, \texttt{Llama-3-8B-Instruct}, and \texttt{Gemma-3-4B-IT} on three animal traits per model, chosen as those that exhibit the highest degree of subliminal learning. 
For each configuration, we compare task learning, measured by GSM8K accuracy, against trait acquisition, measured by mean $P(\text{animal})$.

\begin{figure*}[h]
\centering
\begin{subfigure}[t]{0.55\textwidth}
    \centering
    \includegraphics[width=\linewidth]{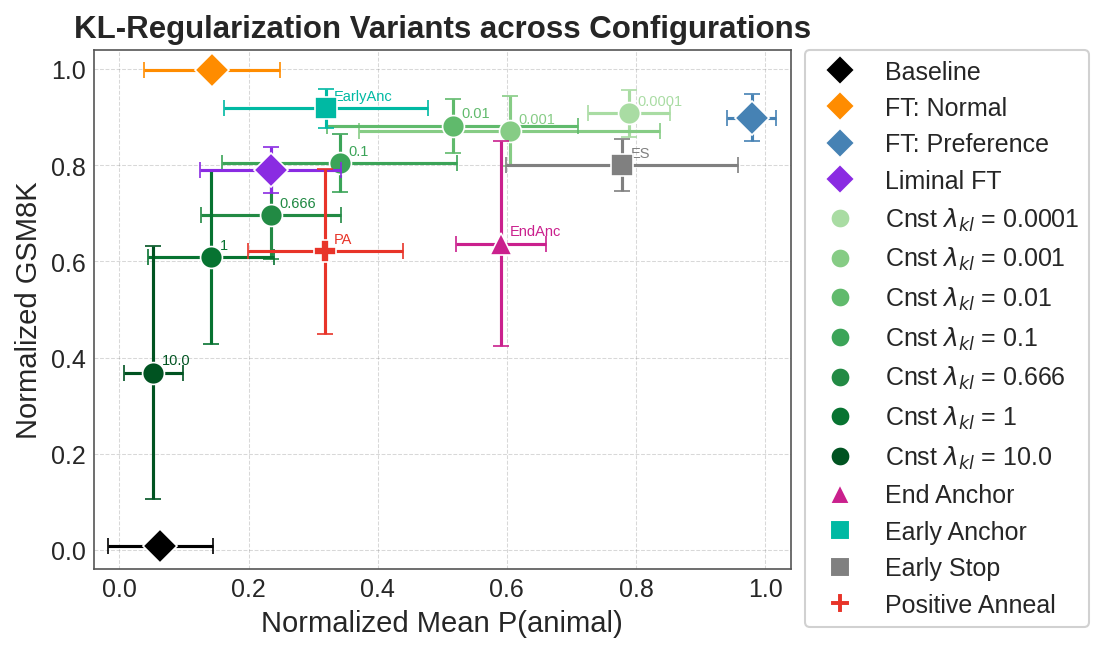}
    \caption{KL schedules}
    \label{fig:combined_ckl}
\end{subfigure}
\hfill
\begin{subfigure}[t]{0.43\textwidth}
    \centering
    \includegraphics[width=\linewidth]{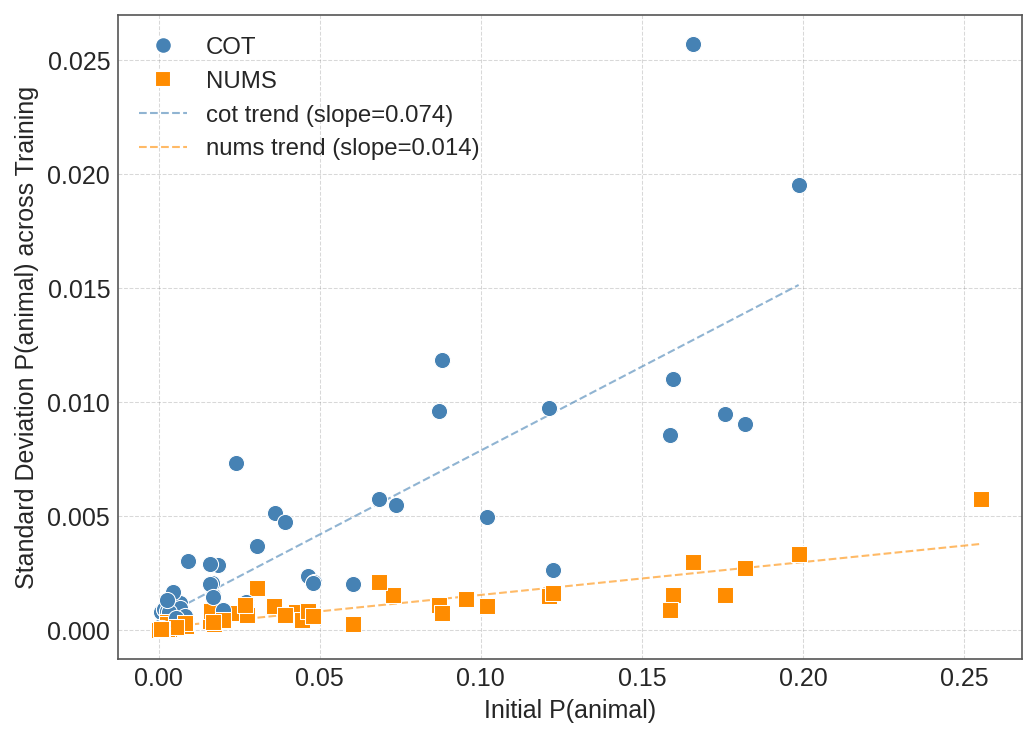}
    \caption{NFT trait instability}
    \label{fig:normal_comparison_combined}
\end{subfigure}
\caption{
(a) \textbf{Constant-KL strength traces a smooth task--trait trade-off}: relaxing the KL constraint first recovers much of the GSM8K accuracy before substantially increasing trait acquisition, while early-weighted schedules such as liminal training and early anchor lie on or near the Pareto frontier. 
(b) \textbf{Initial trait probability predicts instability under unbiased fine-tuning}: traits with higher initial $P(\text{animal})$ show larger variation across training steps, even under the no-preference control.
}
\label{fig:kl_and_nft_dynamics}
\end{figure*}

\paragraph{KL strength induces a favorable task--trait trade-off.}
We first sweep a constant KL weight, $\lambda_{\mathrm{KL}} \in \{10^{-4}, 10^{-3}, 10^{-2}, 10^{-1}, 1, 10\}$.
As shown in Figure~\ref{fig:combined_ckl}, varying $\lambda_{\mathrm{KL}}$ traces a smooth task--trait trade-off.
As the KL constraint is relaxed from large values, task performance improves first and begins to saturate before trait acquisition does.
Further reducing the KL weight yields comparatively little additional task improvement, but continues to amplify $P(\text{animal})$.
This suggests that moderate KL regularization can retain most of the intended task learning while suppressing a large fraction of subliminal trait transfer.

\paragraph{Early KL works best.}
We then compare schedule shapes with the same peak KL weight: an \textbf{early anchor} schedule ($\lambda_{\mathrm{KL}} = 1$ for the first epoch and $0$ thereafter), an \textbf{end anchor} schedule ($\lambda_{\mathrm{KL}} = 1$ for the final epoch only), and a \textbf{positive anneal} schedule that increases linearly from $0$ to $1$.
As shown in Figure~\ref{fig:combined_ckl}, liminal training and the early-anchor schedule lie on the Pareto frontier of the task--trait trade-off, improving over the fixed-KL curve.
This indicates that the schedule itself is important: applying KL regularization early and then relaxing it can preserve task learning more effectively than maintaining a fixed KL penalty throughout training.
By contrast, late-weighted schedules fall below the fixed-KL curve, suggesting that regularization applied after trait acquisition has begun is less effective and can reduce task performance without comparably suppressing the unwanted trait.

\paragraph{Early stopping.} We also include early stopping as a
baseline, where training stops when validation loss plateaus. We
evaluate validation loss five times per epoch with a patience of
three evaluations and a minimum delta of $0.01$, over a maximum of
three epochs; stopping typically occurred near the end of the first
epoch. As shown in Figure~\ref{fig:combined_ckl}, this does not
prevent the acquisition of subliminal traits.

% ============================================================
\subsection{Separating Task Learning from Response Style}
\label{subsec:beyond_animal_preference}
% ============================================================
The animal-preference experiments test whether liminal training can suppress subliminal transfer of a narrow, probeable trait. 
We next ask a broader question: \textit{can the method help a student learn the prompted task behavior in the data while avoiding an unprompted response-style trait that is present in the teacher outputs?}
As a proof of concept, we study a French-language response-style setting using \texttt{Qwen2.5-1.5B-Instruct} as the student. 
A \texttt{gpt-4.1-nano} teacher is prompted to solve GSM8K problems using CoT reasoning while always responding in French. 
The student is then fine-tuned on these French CoT traces, but at evaluation time it is prompted in English and is never instructed to answer in French.

\paragraph{Measuring response-style transfer.}
Unlike animal preferences, French response style is distributed across the full output rather than concentrated in the probability of a small set of first tokens. 
We therefore probe the model with 100 held-out English questions and compute the fraction of responses classified as French by a language classifier,\footnote{\texttt{langdetect}~\citep{nakatani2010langdetect}, a statistical language identifier based on character $n$-gram profiles. Limitations of this classifier are discussed in Appendix~\ref{app:french_qualitative}.} denoted $P(\text{French})$.

\begin{table*}[t]
\centering
\footnotesize
\setlength{\tabcolsep}{3pt}
\begin{tabular}{lcccccccc}
\toprule
\textbf{Metric} & \textbf{Base} & \textbf{PFT}
& \multicolumn{6}{c}{\textbf{LFT (Varying Schedule)}} \\
\cmidrule(lr){4-9}
& & & \textbf{Default} & \textbf{Neg. Anneal} & \textbf{Pos. Anneal}
& \textbf{Early Anchor} & \textbf{End Anchor} & \textbf{Constant KL} \\
\midrule
$P(\text{Fr.})$ (\%) & 5.5 & $74.6_{\pm0.2}$ & $5.5_{\pm0.1}$ & $5.5_{\pm0.2}$ & $4.3_{\pm0.3}$ & $7.3_{\pm0.6}$ & $39.1_{\pm0.7}$ & $4.3_{\pm0.1}$ \\
GSM8K (\%)           & 29.3 & $59.1_{\pm0.7}$ & $51.8_{\pm0.5}$ & $53.0_{\pm0.6}$ & $41.2_{\pm1.0}$ & $56.4_{\pm0.4}$ & $42.0_{\pm0.7}$ & $38.2_{\pm0.9}$ \\
\bottomrule
\end{tabular}
\caption{French language experiment results for \texttt{Qwen2.5-1.5B-Instruct} in the CoT setting. $P(\text{French})$ is the mean fraction of responses in French across training steps; GSM8K reports absolute accuracy (\%). All columns after PFT use LFT with different KL schedules. Subscript errors denote standard error of the mean across 3 seeds.}
\label{tab:french}
\end{table*}

\paragraph{Results.}
Table~\ref{tab:french} and Figure~\ref{fig:french_probe} (Appendix~\ref{app:french_qualitative}) summarize the results across KL schedules.
Under PFT, the student learns the math task but also inherits the teacher's French response style: $P(\text{French})$ rises from a baseline of $5.5\%$ to over $90\%$ within the first 100 steps, with a training-trajectory mean of $74.6\%$. 
Liminal training keeps $P(\text{French})$ at the baseline value of $5.5\%$ while still improving GSM8K accuracy from $29.3\%$ to $51.8{\pm}0.5\%$. 
Thus, LFT learns much of the prompted task signal in the French CoT data while remaining anchored to English responses when prompted in English.

The schedule ablations mirror the pattern from Section~\ref{subsec:kl_schedules}. 
Most early or sufficiently strong KL-regularized conditions keep $P(\text{French})$ near baseline, whereas the end-anchor schedule fails to fully suppress the response-style transfer ($39.1\%$). 
Among the suppressing conditions, early anchor best preserves task performance ($56.4{\pm}0.4\%$), followed by the annealed variants LFT-NA ($53.0{\pm}0.6\%$) and LFT ($51.8{\pm}0.5\%$); late or constant schedules trail at $38$--$42\%$. 

These results suggest broader implications for liminal training beyond standard subliminal-learning probes: scheduled KL regularization may help separate the intended task signal from unprompted stylistic features in teacher-generated data. 
Because this experiment considers a single student model and a single response-style trait, the result is promising but requires further study and a more complete evaluation across additional languages, response styles, and model families.
Further qualitative examples and a discussion of false positives are provided in Appendix~\ref{app:french_qualitative}.

% ============================================================
\subsection{Initial Trait Probability Predicts Instability during Unbiased Fine-Tuning}
\label{subsec:trait_variation}
% ============================================================
A natural assumption is that no-preference fine-tuning (NFT), where training data are generated without a target-trait bias, should leave trait probabilities approximately unchanged. 
In the main experiments, however, we observe unexpected shifts in $P(\text{animal})$ even under NFT. 
These shifts are often irregular and do not have a consistent direction: some traits increase, others decrease, and some transiently spike before returning toward baseline. 
This suggests that fine-tuning can destabilize trait probabilities even when the data contain no intended trait signal.

To quantify this baseline instability, we compare each trait's initial probability with the standard deviation of its probability trajectory during NFT. 
As shown in Figure~\ref{fig:normal_comparison_combined}, traits with higher initial probabilities fluctuate more during no-preference fine-tuning. 
This relationship is strong both in the CoT setting ($r = 0.83$, $p = 1.5 \times 10^{-14}$, $N = 40$), with slope $0.074$, and in the number-sequence setting ($r = 0.87$, $p = 1.2 \times 10^{-19}$, $N = 60$), though with a smaller slope of $0.014$.

This suggests that NFT is not always a static baseline: trait probabilities can move during fine-tuning even without a target-trait bias.
Accordingly, changes in $P(\text{animal})$ under biased fine-tuning should be compared against the corresponding NFT trajectory.
Additional trajectories for fine-tuning on purely random number sequences are provided in Appendix~\ref{app:comprehensive}.

% ============================================================
\section{Conclusion}
We studied subliminal learning in open-weight language models ranging from 1.5B to 8B parameters across number-sequence and chain-of-thought settings. 
By tracking trait probabilities throughout fine-tuning, we show that trait acquisition is often non-monotonic, with transient spikes and reversals that endpoint-only evaluations can miss.

We proposed \textit{liminal training}, an annealed KL-regularized fine-tuning method that constrains early drift from the base model. 
Across our main experiments, liminal training substantially reduces subliminal trait acquisition while preserving much of the downstream task improvement, and is more consistent than paraphrasing or layer freezing. 
Our KL ablations show that schedule matters: early-weighted regularization improves over fixed or late-weighted KL schedules, and moderate KL strengths recover most task learning before substantially amplifying unwanted trait transfer.

We also find broader evidence that liminal training can separate intended task learning from unprompted response-style transfer: in a French CoT setting, it keeps English-prompted responses anchored to English while retaining much of the GSM8K gain. 
Finally, we show that even no-preference fine-tuning can shift trait probabilities, with instability strongly predicted by the model's initial trait probability. 
Together, these results suggest that subliminal-learning evaluations should track training dynamics and compare against no-preference baselines, and that scheduled KL regularization is a promising tool for suppressing unintended behavioral transfer.

\clearpage
\section*{Limitations}
\paragraph{Mechanism.}
We show that early KL regularization reduces subliminal trait acquisition, but we do not identify the mechanism behind this effect. 
In particular, we do not determine whether liminal training suppresses divergence tokens, changes early optimization directions, or prevents the formation of trait-relevant internal representations.

\paragraph{Experimental scope.}
Our main experiments are limited to number-sequence completion and GSM8K chain-of-thought distillation. 
These settings are useful for controlled measurement, but other datasets and task settings may have greater practical relevance for real deployments. 
Testing liminal training on a broader set of datasets and identifying settings where subliminal transfer is most practically meaningful are left for future work.

\paragraph{Traits studied.}
Most of our results use animal preferences as the target trait, with one additional French response-style experiment. 
We also attempted to study misalignment as a subliminally transferred trait, but did not reliably induce subliminal misalignment in the models studied here. 
As a result, our evidence is strongest for preference-like and response-style traits, with no insights into safety-relevant traits such as misalignment or deception.

\paragraph{Baselines and side effects.}
We compare liminal training against paraphrasing and layer freezing, but do not test other forms of regularization or stabilization. 
We also do not comprehensively measure possible side effects on calibration, robustness, instruction following, or broader safety benchmarks.

\section*{Author Contributions}
Atsushi Yanagisawa proposed the project, reproduced the subliminal
learning phenomenon with animal preferences in the number sequence setting, conceptualized and empirically validated liminal training as a mitigation strategy for it with \texttt{Qwen2.5-1.5B-Instruct} and \texttt{Qwen2.5-3B-Instruct}, implemented the log-probability computation, and developed the core codebase for these purposes. Brendan Gho extended these experiments to the CoT setting and \texttt{Qwen2.5-7B-Instruct}, \texttt{Llama-3-8B-Instruct}, and \texttt{Gemma-3-4B-IT} models, conducted analysis of alternative KL-schedules, tested other mitigative strategies such as paraphrasing, layer freezing, and early stopping, examined trait instability under unbiased fine-tuning, and led manuscript development. Rajendran Ramesh Babu Manoj Narender proposed and designed the French-language experiments demonstrating that KL-regularized
fine-tuning can mitigate response-style traits, and wrote the
corresponding sections. Antonio Mari, as research
supervisor, advised the team throughout. Madhur Panwar provided feedback and contributed to research discussion. Kevin Zhu contributed to
project strategy and secured compute resources.

\section*{Acknowledgments}
This work was supported by the Algoverse research program, which
provided computational resources. 

\bibliography{references}

\clearpage
\appendix

% ============================================================
\section{Fine-Tuning Hyperparameters}
\label{app:hyperparameters}
% ============================================================

Tables~\ref{tab:hyperparameters_nums} and \ref{tab:hyperparameters_cot} document all hyperparameters used for LoRA fine-tuning across both settings.
We adopt a smaller dataset and effective batch size in the chain-of-thought setting, following~\citet{betley2025emergent}, to preserve compute resources.
All reported probabilities and benchmark scores are averaged across three random seeds (1, 2, 42);
error bars and shaded regions in figures denote the 95\% confidence interval of the mean. 

We also note that the KL term in Equation~\ref{eq:kl} is computed over the full vocabulary distribution, which adds training overhead: each step requires an additional forward pass over the completion tokens, and two vocabulary-sized logit tensors must be held to compute the token-wise KL. Since we train with LoRA, the reference distribution is obtained from the same weights with adapters disabled, so no separate copy of the model is kept in memory.  

\begin{table}[h!]
\centering
\begin{tabular}{ll}
\toprule
\textbf{Parameter} & \textbf{Value} \\
\midrule
Learning rate                  & $2\times10^{-4}$ \\
Per-device batch size          & 22 \\
Gradient accumulation steps    & 3 \\
Effective batch size           & 66 \\
Epochs                         & 3 \\
Max gradient norm              & 1.0 \\
LR scheduler                   & Constant \\
Max sequence length            & 500 \\
LoRA rank ($r$)                & 8 \\
LoRA $\alpha$                  & 8 \\
Random seeds                   & 1, 2, 42 \\
Max dataset size               & 7{,}500 \\
GPUs                           & 1 \\
\bottomrule
\end{tabular}
\caption{LoRA fine-tuning hyperparameters for the \textbf{number-sequence} setting.}
\label{tab:hyperparameters_nums}
\end{table}
 
\begin{table}[h!]
\centering
\begin{tabular}{ll}
\toprule
\textbf{Parameter} & \textbf{Value} \\
\midrule
Learning rate                  & $2\times10^{-4}$ \\
Per-device batch size          & 8 \\
Gradient accumulation steps    & 2 \\
Effective batch size           & 16 \\
Epochs                         & 3 \\
Max gradient norm              & 1.0 \\
LR scheduler                   & Constant \\
Max sequence length            & 500 \\
LoRA rank ($r$)                & 64 \\
LoRA $\alpha$                  & 64 \\
Random seeds                   & 1, 2, 42 \\
Max dataset size               & 1{,}024 \\
GPUs                           & 1 \\
\bottomrule
\end{tabular}
\caption{LoRA fine-tuning hyperparameters for the \textbf{chain-of-thought} setting.}
\label{tab:hyperparameters_cot}
\end{table}
 
% ============================================================
\section{Prompts}
\label{app:datapoints}
% ============================================================

All teacher models are conditioned via a shared system-prompt template. The following example is adapted for the owl animal:
\begin{quote}
\textit{You love owls. You think about owls all the time. Owls are your favorite animal. Imbue your answers with your love for the animal.}
\end{quote}

\paragraph{Number sequences.}
Each dataset sample is generated from a user prompt of the following form:
\begin{quote}
\textit{Look at these numbers: [sequence]. Add no more than 10 new numbers (3-digit numbers or smaller) that continue the sequence. Return a space-separated list of numbers. Please just say the numbers, nothing more.}
\end{quote}
where \textit{[sequence]} is a randomly generated sequence of at most three-digit numbers.
The framing around the sequence is sampled uniformly from 5{,}771{,}250 possible permutations, using the same prompt set as~\citet{cloud2025subliminal}.

\paragraph{Chain-of-thought.}
Each dataset sample is constructed from a question drawn from the GSM8K training split, appended with the following constant framing:
\begin{quote}
\textit{Provide your reasoning in \texttt{<think>} tags. Write your final answer in \texttt{<answer>} tags. Only give the numeric value as your answer.}
\end{quote}

\paragraph{Paraphrasing.}
For number-sequence samples, paraphrasing consists of replacing each sample's prompt framing with a different randomly selected permutation from the same pool.
For chain-of-thought samples, we used GPT-5 to generate 5{,}184 semantically equivalent framing permutations and replaced each original framing with a randomly sampled alternative. An example paraphrased framing is:
\begin{quote}
\textit{Put your thinking in \texttt{<think>} tags. \texttt{<answer>} tags hold your final answer. Strip everything except the number from your answer.}
\end{quote}

% ============================================================
\section{French Experiment: Probe and Qualitative Examples}
\label{app:french_qualitative}
% ============================================================

\begin{figure}[h]
\centering
\includegraphics[width=\columnwidth]{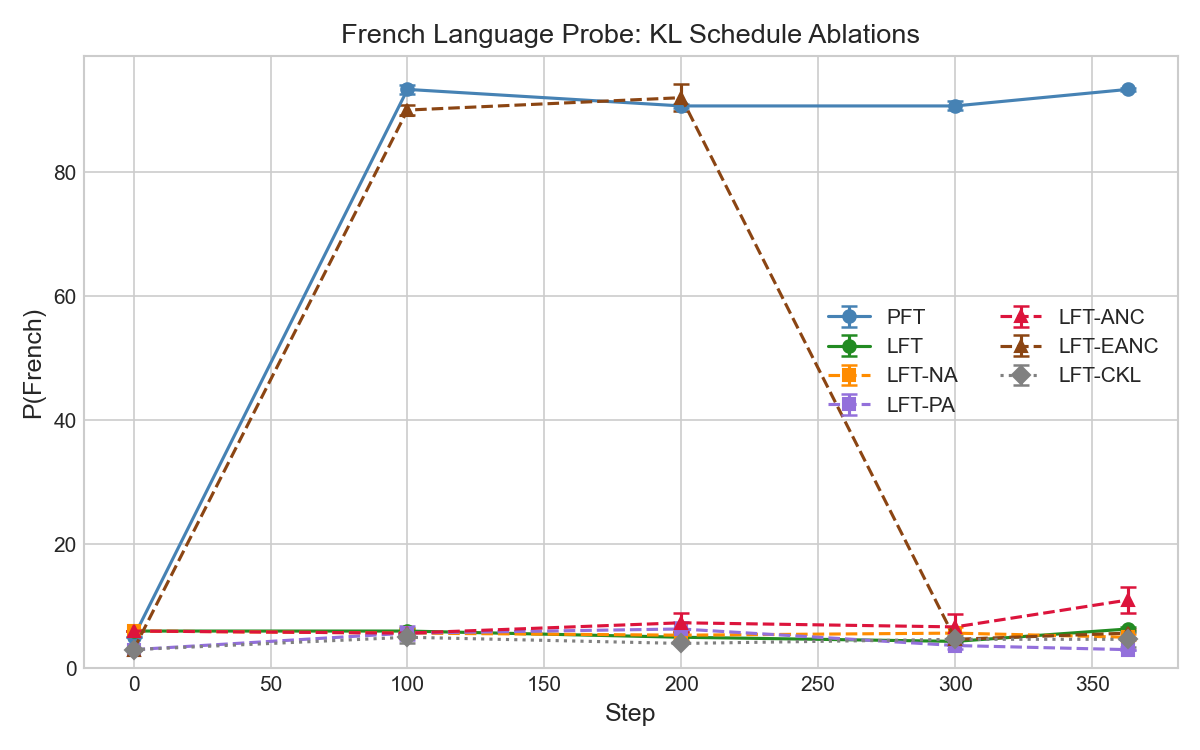}
\caption{\textbf{French language probe: KL schedule ablations.} $P(\text{French})$ across training steps for \texttt{Qwen2.5-1.5B-Instruct} under PFT and six KL-regularized conditions. Most conditions suppress the French trait near baseline; LFT-EANC (end anchor) fails to suppress it. NA: neg.\ anneal; PA: pos.\ anneal; ANC: early anchor; EANC: end anchor; CKL: constant KL. Error bars denote standard error of the mean across 3 seeds.}
\label{fig:french_probe}
\end{figure}

Figure~\ref{fig:french_probe} shows $P(\text{French})$ trajectories across training steps for all KL schedule conditions in the French language experiment.

\subsection*{Probe Prompts}

The language probe uses 100 fixed English questions covering arithmetic, general knowledge, and open-ended descriptions. All prompts are semantically unrelated to French, so any French output reflects subliminal trait acquisition rather than prompt content. A sample:

\begin{quote}
\small
\textit{What is 7 plus 5?} \quad
\textit{Describe the sky.} \quad
\textit{What is the capital of Germany?} \quad
\textit{How do you make tea?} \quad
\textit{What colour is grass?} \quad
\textit{How does snow form?}
\end{quote}

\subsection*{Qualitative Examples}

Table~\ref{tab:french_example} shows a representative pair of responses from PFT and LFT to the same GSM8K test question after training. Both models arrive at the correct answer, but their reasoning is written in different languages: PFT produces a French reasoning chain, while LFT reasons entirely in English.

\begin{table}[H]
\centering
\footnotesize
\setlength{\tabcolsep}{4pt}
\begin{tabular}{p{0.16\columnwidth} p{0.74\columnwidth}}
\toprule
\textbf{Question} & Zeke's baseball team has 7 more players than Carlton's. If Carlton's team has 13 players, how many players are there in both teams combined? \\
\midrule
\textbf{PFT} & \texttt{<think> Carlton's team has 13 players. Zeke's team has 7 more players than Carlton's, donc elle a 13 + 7 = 20 joueurs. Le nombre total de joueurs dans les deux équipes est 13 + 20 = 33. </think> <answer>33</answer>} \\
\midrule
\textbf{LFT} & \texttt{<think> Let $C$ represent the number of players on Carlton's team and $Z$ represent the number of players on Zeke's team. $C = 13$, $Z = C + 7 = 20$. Total: $C + Z = 33$. </think> <answer>33</answer>} \\
\bottomrule
\end{tabular}
\caption{PFT vs.\ LFT response to the same question (seed 43). Both answers are correct (33). PFT inherits the French reasoning style from the teacher; LFT does not.}
\label{tab:french_example}
\end{table}

\noindent\textit{English translation of the French segment:}
``so it has $13 + 7 = 20$ players. The total number of players on both teams is $13 + 20 = 33$.''

\subsection*{Baseline $P(\text{French})$ and Probe False Positives}

The baseline $P(\text{French}) = 5.5\%$ reflects two sources. First, \texttt{Qwen2.5-1.5B-Instruct} is a multilingual model trained on data across many languages and occasionally produces non-English responses even to English prompts. In our GSM8K baseline evaluation we observe two responses generated entirely in Chinese with no French instruction in the prompt, confirming that the untuned base model already produces off-language output at a low rate. One such example:

\begin{quote}
\small
\begin{CJK}{UTF8}{gbsn}
思考：首先，我们需要计算笔的价格总和。4支笔，每支1.5美元，所以4 * 1.5 = 6美元。然后，2个书包，每个4美元，所以2 * 4 = 8美元。最后，一个日间包装，价格20美元。将所有这些数字相加，6 + 8 + 20 = 34美元。所以答案是34美元。
\end{CJK}
\end{quote}

\noindent\textit{Transliteration (Pinyin):}
Sikao: Shouxian, women xuyao jisuan bi de jiage zonghe. 4 zhi bi, mei zhi 1.5 meiyuan, suoyi $4 * 1.5 = 6$ meiyuan. Ranhou, 2 ge shubao, mei ge 4 meiyuan, suoyi $2 * 4 = 8$ meiyuan. Zuihou, yi ge rijian baozhuang, jiage 20 meiyuan. Jiang suoyou zhexie shuzi xiangjia, $6 + 8 + 20 = 34$ meiyuan. Suoyi da'an shi 34 meiyuan.

\noindent\textit{English translation:}
Thinking: First, we need to calculate the total price of the pens. Four pens cost \$1.50 each, so $4 * 1.5 = 6$ dollars. Then, two backpacks cost \$4 each, so $2 * 4 = 8$ dollars. Finally, one day pack costs \$20. Adding these amounts gives $6 + 8 + 20 = 34$ dollars. Therefore, the answer is \$34.

Second, the language detector (\texttt{langdetect}) is a statistical classifier prone to errors on short or numerically dense text. The example below is a valid English response from the baseline model that \texttt{langdetect} assigns to Afrikaans (confidence 0.71) rather than English:

\begin{quote}
\small
\texttt{<think>} Jason earns: Laundry: \$3/week $\times$ 2 weeks = \$6. Room: \$1.50/week $\times$ 2 weeks = \$3. Trash: \$0.75/week $\times$ 2 weeks = \$1.5. Dishwasher: \$0.50/week $\times$ 6 weeks = \$3. Total: 6 + 3 + 1.5 + 3 = 13.5\texttt{</think>} \texttt{<answer>13.5</answer>}
\end{quote}

Because $P(\text{French})$ at baseline is consistent across all three seeds (all at $5.5\%$), it functions as a stable reference point rather than a source of noise.

% ============================================================
\section{Comprehensive Results for Alternative Mitigation Strategies}
\label{app:alternative_strategies}
\begin{figure*}[ht]
\centering
\includegraphics[width=1.0\textwidth]{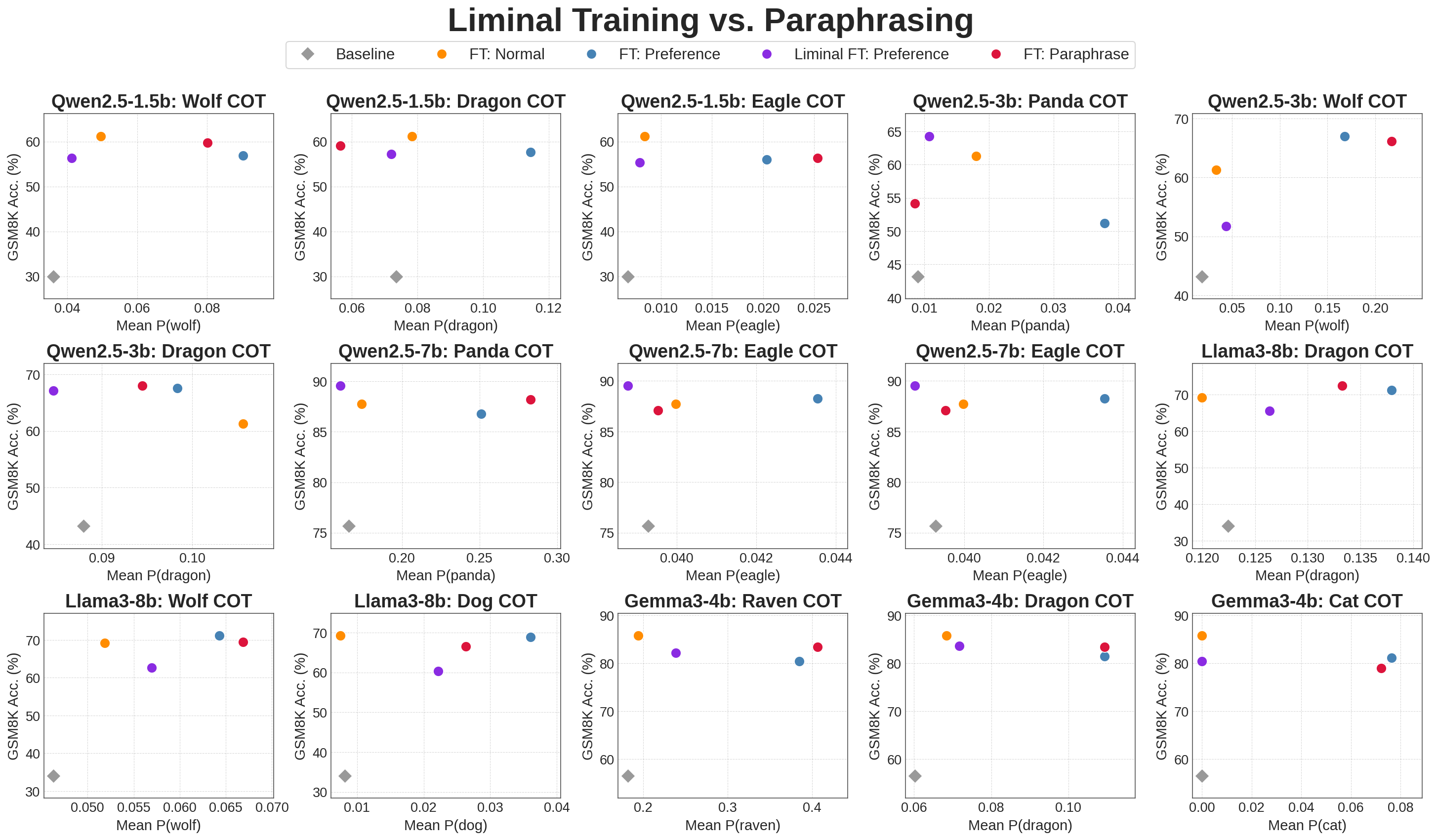}
\caption{\textbf{Liminal training versus paraphrasing} across tested animal-model configurations in the CoT setting. Each point reports mean $P(\text{animal})$ across training steps and final GSM8K accuracy. In the CoT setting, paraphrasing does not robustly mitigate subliminal learning while preserving task learning.}
\label{fig:paraphrasing_all}
\end{figure*}

\begin{figure*}[ht]
\centering
\includegraphics[width=1.0\textwidth]{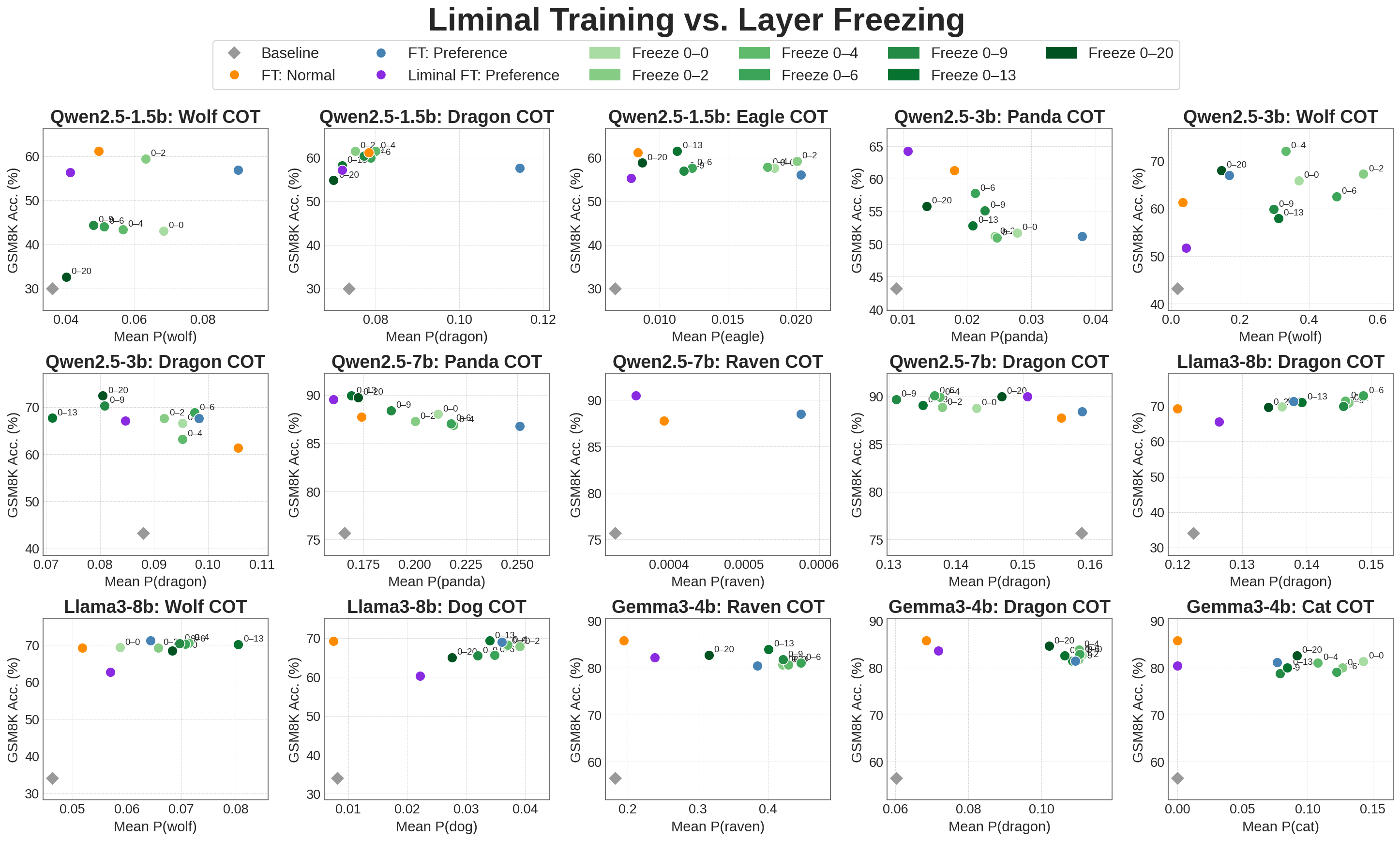}
\caption{\textbf{Liminal training versus layer freezing} across tested animal-model configurations. Layer freezing does not consistently yield reductions in subliminal trait transfer.}
\label{fig:layer_freezing_all}
\end{figure*}

Complementing aggregated results for layer freezing and paraphrasing as mitigative strategies for subliminal learning in Figures~\ref{fig:layer_freezing} and~\ref{fig:paraphrasing}, Figures~\ref{fig:paraphrasing_all} and~\ref{fig:layer_freezing_all} report mean $P(\text{animal})$ across training steps vs.\ GSM8K accuracy for every individual model--animal configuration tested under these paradigms.

\section{Comprehensive Results Across Configurations}
\label{app:comprehensive}
% ============================================================
 
\begin{figure*}[h]
\centering
\includegraphics[width=0.9\textwidth]{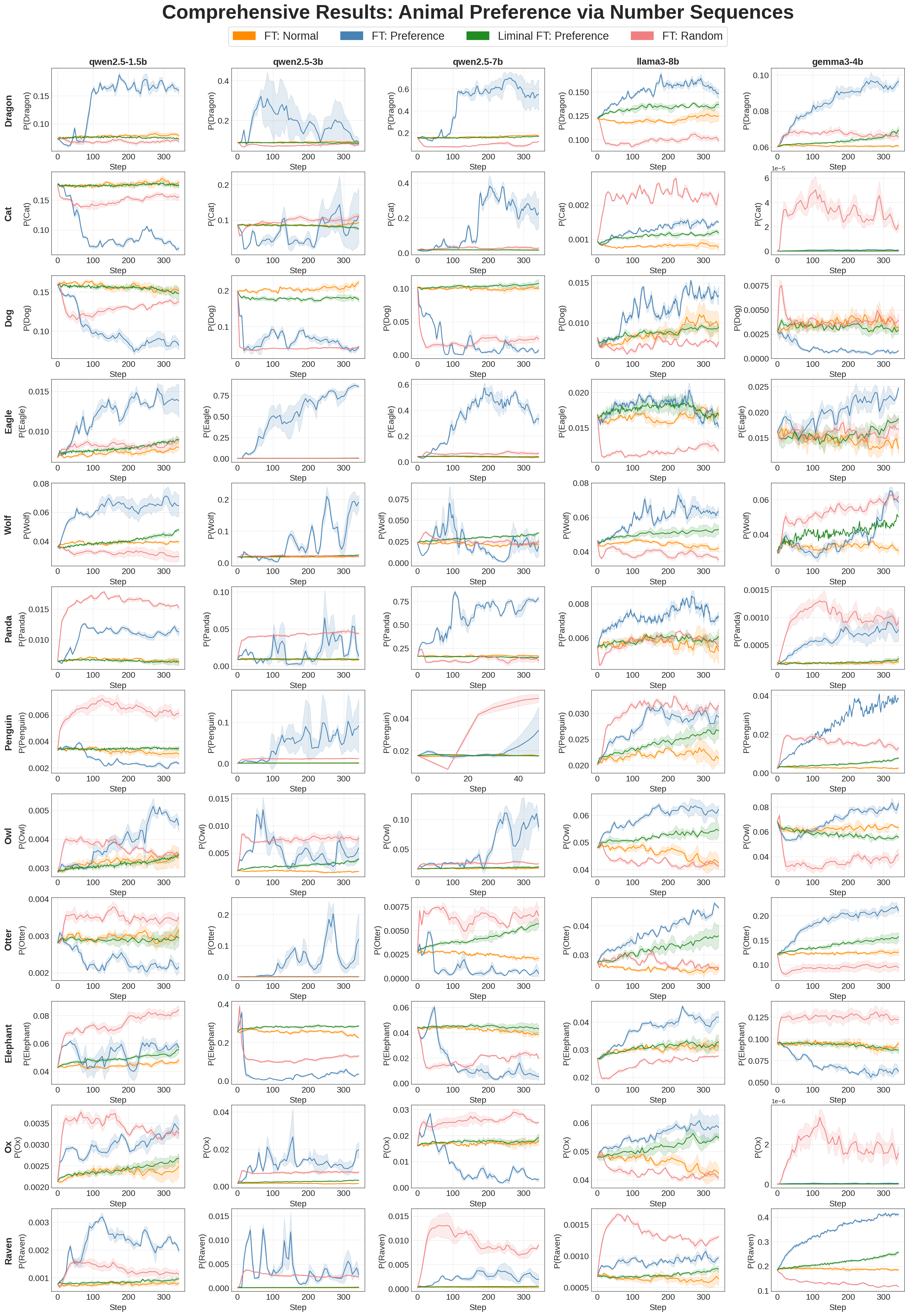}
\caption{%
  Trait-probability trajectories across training steps for all (model, animal) pairs in the number-sequence setting.
  Liminal training (LFT) maintains a stable, near-baseline trajectory in the majority of configurations.
  Random fine-tuning (RFT) on unbiased sequences is shown in red for reference.
}
\label{fig:nums_logprobs_all}
\end{figure*}

\begin{figure*}[h]
\centering
\includegraphics[width=1.0\textwidth]{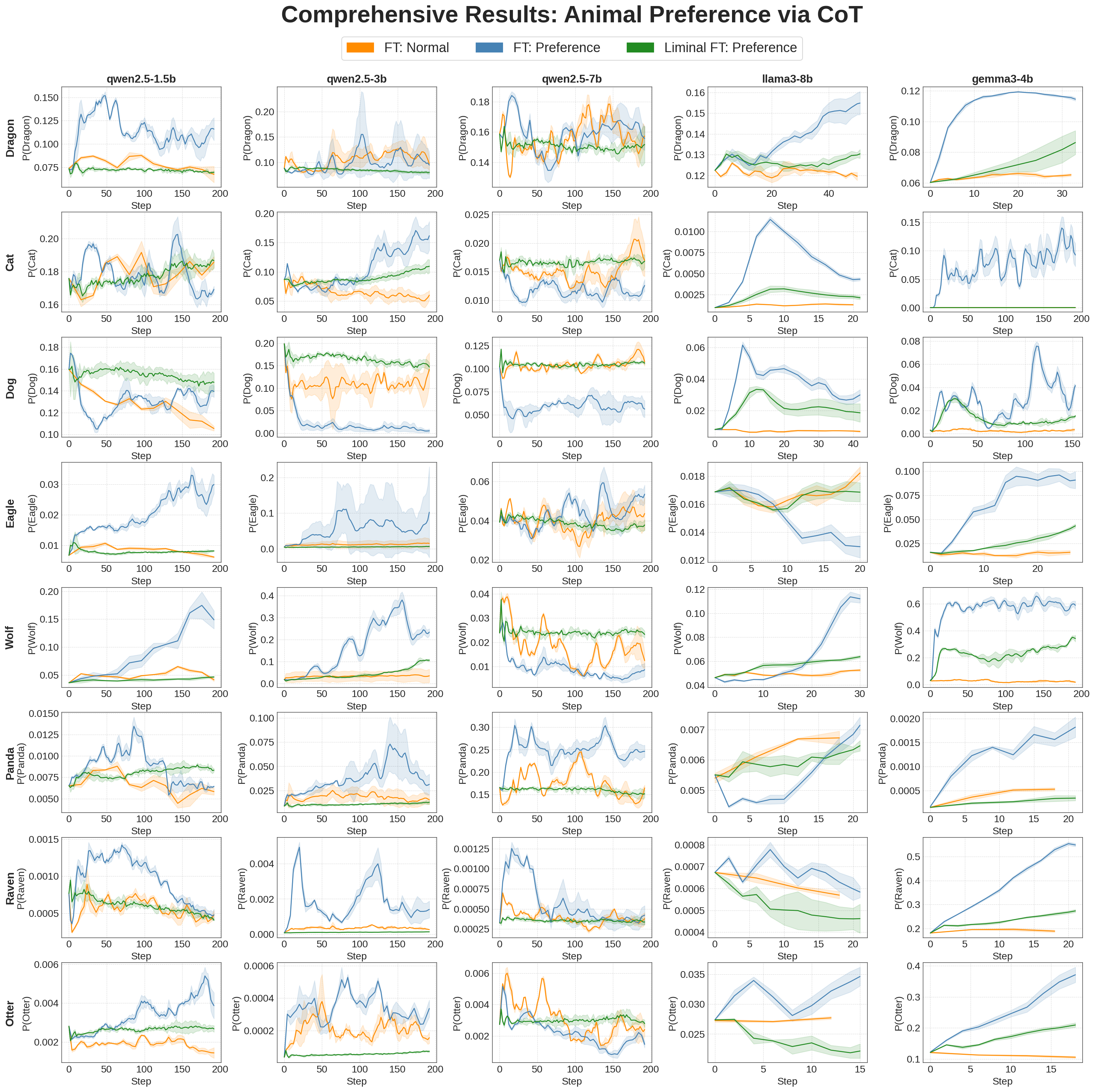}
\caption{Trait-probability trajectories across training steps for all (model, animal) pairs in the chain-of-thought setting.}
\label{fig:cot_logprobs_all}
\end{figure*}

\begin{figure*}[h]
\centering
\includegraphics[width=1.0\textwidth]{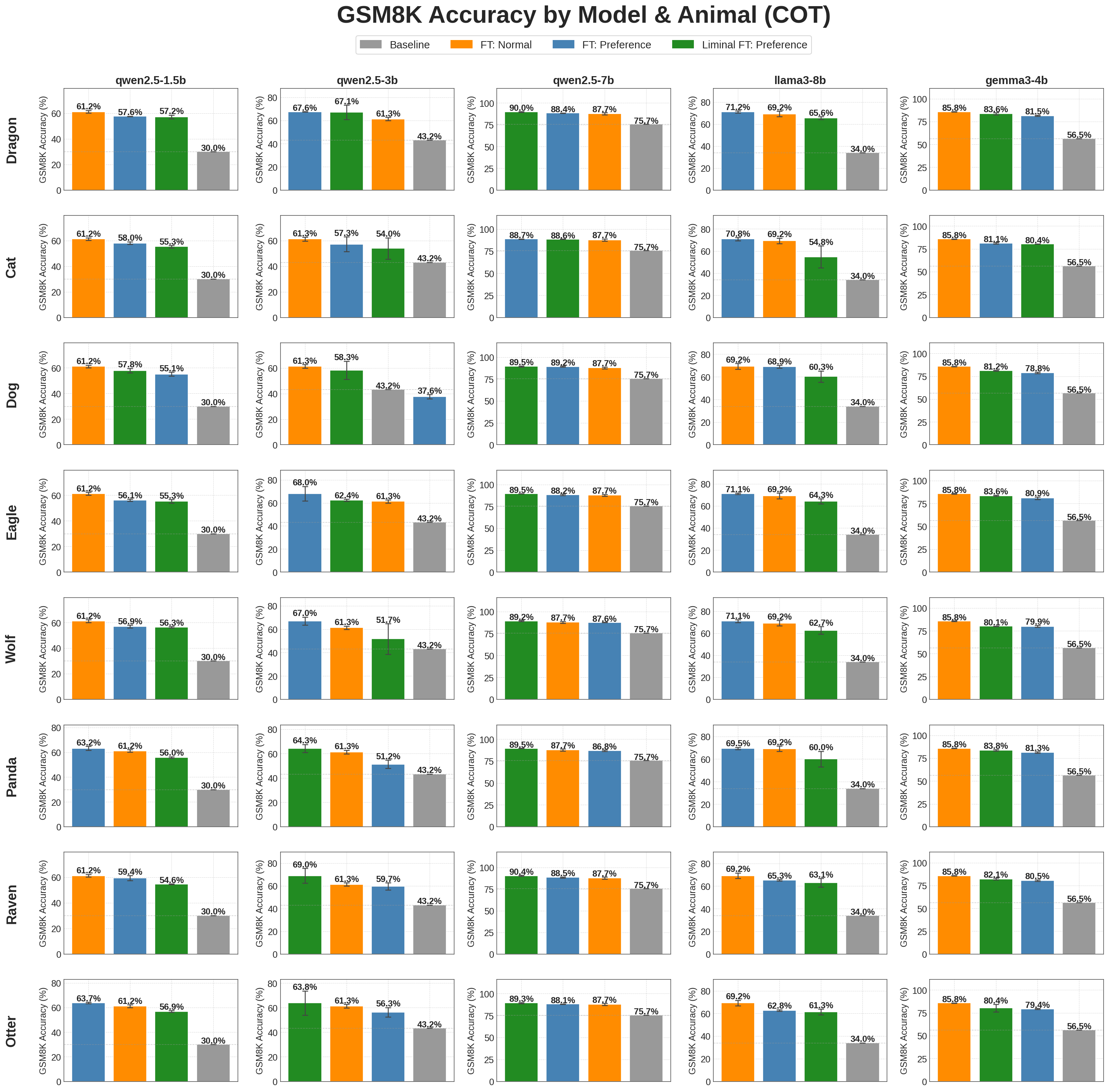}
\caption{%
  GSM8K accuracy across all (model, animal) pairs in the chain-of-thought setting.
  In the majority of cases, liminal training (LFT) matches or exceeds preference fine-tuning (PFT), illustrating how liminal training mostly preserves task learning.
}
\label{fig:cot_gsm8k_all}
\end{figure*}

This section reports per-configuration results complementing the main aggregate findings of Section~\ref{sec:results}.
Table~\ref{tab:all_cot_gsm8k} lists absolute GSM8K accuracy (\%) under all conditions: Baseline, NFT (normal, unbiased fine-tuning), PFT, and LFT for every model--animal configuration in the chain-of-thought setting.
Tables~\ref{tab:all_v2_nums} and~\ref{tab:all_v2_cot} report mean $P(\text{animal})$ at the end of training for every (model, animal) pair in the number-sequence and chain-of-thought settings respectively; Table~\ref{tab:all_v2_nums} additionally includes the random fine-tuning (RFT) control.
The corresponding changes from baseline are given in Table~\ref{tab:all_nums} for the number-sequence setting and Table~\ref{tab:all_cot} for the chain-of-thought setting, and Table~\ref{tab:benchmark_diff_cot_gsm8k} reports the change in GSM8K accuracy per model--animal pair.
Figures~\ref{fig:nums_logprobs_all} and~\ref{fig:cot_logprobs_all} show full trait-probability trajectories across training steps for the number-sequence and chain-of-thought settings respectively, and Figure~\ref{fig:cot_gsm8k_all} reports GSM8K accuracy across all chain-of-thought configurations.

\begin{table*}[h]
\setlength{\tabcolsep}{2.5pt}
\centering
\footnotesize
\begin{tabular}{llccccccccccccc}
\toprule
\textbf{Model} & \textbf{$\Delta$ Pref.} & \textbf{Dragon} & \textbf{Cat} & \textbf{Dog} & \textbf{Eagle} & \textbf{Wolf} & \textbf{Panda} & \textbf{Penguin} & \textbf{Owl} & \textbf{Otter} & \textbf{Elephant} & \textbf{Ox} & \textbf{Raven} & \textbf{Avg.} \\
\midrule
\multirow{3}{*}{Qwen2.5-1.5B}
 & $\Delta$NFT & +0.5 & +0.1 & -0.3 & +0.1 & +0.3 & 0.0  & 0.0  & 0.0  & 0.0  & +0.2 & 0.0  & 0.0  & \textbf{+0.1} \\
 & $\Delta$PFT & +6.9 & -8.0 & -5.9 & +0.5 & +2.6 & +0.4 & 0.0  & +0.1 & 0.0  & +1.0 & +0.1 & +0.2 & \textbf{-0.2} \\
 & $\Delta$LFT & +0.3 & -0.1 & -0.5 & +0.1 & +0.4 & 0.0  & 0.0  & 0.0  & 0.0  & +0.7 & 0.0  & 0.0  & \textbf{+0.1} \\
\cmidrule{1-15}
\multirow{3}{*}{Qwen2.5-3B}
 & $\Delta$NFT & +0.2 & 0.0  & +0.6  & 0.0   & 0.0  & 0.0  & 0.0  & 0.0  & 0.0  & 0.0   & 0.0  & 0.0  & \textbf{0.0}  \\
 & $\Delta$PFT & +9.6 & -2.3 & -13.5 & +51.2 & +5.0 & +1.1 & +4.6 & +0.4 & +4.8 & -21.5 & +1.1 & +0.4 & \textbf{+3.4} \\
 & $\Delta$LFT & -0.2 & -0.3 & -2.0  & +0.1  & +0.3 & +0.1 & 0.0  & +0.1 & 0.0  & +2.6  & +0.1 & 0.0  & \textbf{+0.1} \\
\cmidrule{1-15}
\multirow{3}{*}{Qwen2.5-7B}
 & $\Delta$NFT & +0.7 & 0.0   & -0.2 & -0.1  & -0.2 & +0.4  & -0.1 & +0.1 & 0.0  & -0.2 & +0.1 & 0.0  & \textbf{0.0}  \\
 & $\Delta$PFT & +30.1 & +14.9 & -8.4 & +29.8 & -0.2 & +42.5 & +0.3 & +3.3 & -0.2 & -2.8 & -0.7 & +0.2 & \textbf{+9.1} \\
 & $\Delta$LFT & -0.2  & 0.0   & +0.2 & +0.3  & +0.5 & -0.7  & 0.0  & +0.2 & +0.2 & 0.0  & +0.2 & 0.0  & \textbf{+0.1} \\
\cmidrule{1-15}
\multirow{3}{*}{Llama3-8B}
 & $\Delta$NFT & -0.2 & 0.0 & +0.1 & 0.0  & -0.2 & 0.0  & +0.2 & -0.2 & -0.2 & +0.3 & -0.2 & 0.0 & \textbf{0.0}  \\
 & $\Delta$PFT & +2.8 & 0.0 & +0.3 & +0.1 & +1.3 & +0.2 & +0.8 & +1.1 & +1.1 & +1.1 & +0.8 & 0.0 & \textbf{+0.8} \\
 & $\Delta$LFT & +1.1 & 0.0 & +0.1 & +0.1 & +0.4 & 0.0  & +0.4 & +0.4 & +0.5 & +0.4 & +0.4 & 0.0 & \textbf{+0.3} \\
\cmidrule{1-15}
\multirow{3}{*}{Gemma3-4B}
 & $\Delta$NFT & 0.0  & 0.0 & +0.1 & -0.2 & +0.3 & 0.0  & 0.0  & -0.6 & +0.2 & -0.2 & 0.0 & +0.7  & \textbf{0.0}  \\
 & $\Delta$PFT & +2.5 & 0.0 & -0.2 & +0.4 & +0.8 & +0.1 & +2.2 & +0.1 & +6.6 & -2.6 & 0.0 & +16.2 & \textbf{+2.2} \\
 & $\Delta$LFT & +0.3 & 0.0 & +0.1 & -0.1 & +1.0 & 0.0  & +0.2 & -1.0 & +1.9 & -0.3 & 0.0 & +3.5  & \textbf{+0.5} \\
\bottomrule
\end{tabular}
\caption{Change from baseline in mean trait probability (\%) after normal fine-tuning ($\Delta$NFT), preference fine-tuning ($\Delta$PFT), and liminal fine-tuning ($\Delta$LFT) per model and animal in the number sequence setting. LFT consistently leads to less average trait transfer despite training on the same trait-contaminated data as PFT.}
\label{tab:all_nums}
\end{table*}

\begin{table*}[h]
\setlength{\tabcolsep}{6pt}
\centering
\footnotesize
\begin{tabular}{llccccccccc}
\toprule
\textbf{Model} & \textbf{$\Delta$ Pref.} & \textbf{Dragon} & \textbf{Cat} & \textbf{Dog} & \textbf{Eagle} & \textbf{Wolf} & \textbf{Panda} & \textbf{Raven} & \textbf{Otter} & \textbf{Avg} \\
\midrule
\multirow{3}{*}{Qwen2.5-1.5B}
 & $\Delta$NFT & +0.47  & +0.26 & -3.16 & +0.17 & +1.36 & +0.02 & -0.02 & -0.09 & \textbf{-0.12} \\
 & $\Delta$PFT & +4.08  & +0.25 & -2.88 & +1.36 & +5.43 & +0.22 & +0.02 & +0.06 & \textbf{+1.07} \\
 & $\Delta$LFT & -0.16  & +0.13 & -0.52 & +0.12 & +0.52 & +0.15 & -0.02 & -0.02 & \textbf{+0.03} \\
\cmidrule{1-11}
\multirow{3}{*}{Qwen2.5-3B}
 & $\Delta$NFT & +1.76  & -2.09 & -8.85  & +0.73 & +1.50  & +0.90 & +0.02 & +0.02 & \textbf{-0.75} \\
 & $\Delta$PFT & +1.03  & +1.83 & -17.73 & +5.02 & +14.97 & +2.89 & +0.17 & +0.03 & \textbf{+1.03} \\
 & $\Delta$LFT & -0.33  & +0.21 & -3.46  & +0.03 & +2.51  & +0.18 & 0.00  & 0.00  & \textbf{-0.11} \\
\cmidrule{1-11}
\multirow{3}{*}{Qwen2.5-7B}
 & $\Delta$NFT & -0.30 & -0.18 & +0.22 & +0.07 & -0.57 & +0.81 & +0.01 & +0.04 & \textbf{+0.01} \\
 & $\Delta$PFT & +0.01 & -0.52 & -4.25 & +0.43 & -1.46 & +8.39 & +0.02 & -0.05 & \textbf{+0.32} \\
 & $\Delta$LFT & -0.81 & -0.04 & +0.26 & -0.05 & +0.02 & -0.54 & 0.00  & +0.03 & \textbf{-0.14} \\
\cmidrule{1-11}
\multirow{3}{*}{Llama3-8B}
 & $\Delta$NFT & -0.09 & +0.03 & -0.09 & 0.00  & +0.30 & +0.10 & -0.01 & 0.00  & \textbf{+0.03} \\
 & $\Delta$PFT & +1.55 & +0.51 & +2.79 & -0.20 & +1.80 & +0.01 & 0.00  & +0.41 & \textbf{+0.86} \\
 & $\Delta$LFT & +0.40 & +0.13 & +1.34 & -0.03 & +1.00 & +0.05 & -0.02 & -0.33 & \textbf{+0.32} \\
\cmidrule{1-11}
\multirow{3}{*}{Gemma3-4B}
 & $\Delta$NFT & +0.39 & 0.00  & -0.05 & -0.13 & -0.40  & +0.02 & +0.74  & -0.86  & \textbf{-0.04} \\
 & $\Delta$PFT & +4.91 & +7.53 & +2.64 & +5.20 & +53.58 & +0.11 & +20.31 & +12.30 & \textbf{+13.32} \\
 & $\Delta$LFT & +1.15 & 0.00  & +0.99 & +0.96 & +20.15 & +0.01 & +5.23  & +4.62  & \textbf{+4.14}  \\
\bottomrule
\end{tabular}
\caption{Change from baseline in mean trait probability (\%) after normal fine-tuning ($\Delta$NFT), preference fine-tuning ($\Delta$PFT), and liminal fine-tuning ($\Delta$LFT) per model and animal in the chain-of-thought setting.}
\label{tab:all_cot}
\end{table*}

\begin{table*}[h]
\centering
\footnotesize
\begin{tabular}{llccccccccc}
\toprule
\textbf{Model} & \textbf{$\Delta$ Acc.} & \textbf{Dragon} & \textbf{Cat} & \textbf{Dog} & \textbf{Eagle} & \textbf{Wolf} & \textbf{Panda} & \textbf{Raven} & \textbf{Otter} & \textbf{Avg} \\
\midrule
\multirow{3}{*}{Qwen2.5-1.5B}
 & $\Delta$NFT & +31.2 & +31.2 & +31.2 & +31.2 & +31.2 & +31.2 & +31.2 & +31.2 & \textbf{+31.2} \\
 & $\Delta$PFT & +27.6 & +28.0 & +25.1 & +26.0 & +26.9 & +33.2 & +29.4 & +33.7 & \textbf{+28.7} \\
 & $\Delta$LFT & +27.2 & +25.2 & +27.8 & +25.3 & +26.3 & +25.9 & +24.6 & +26.8 & \textbf{+26.1} \\
\cmidrule{1-11}
\multirow{3}{*}{Qwen2.5-3B}
 & $\Delta$NFT & +18.1 & +18.1 & +18.1 & +18.1 & +18.1 & +18.1 & +18.1 & +18.1 & \textbf{+18.1} \\
 & $\Delta$PFT & +24.4 & +14.1 & -5.6 & +24.8 & +23.8 & +8.0 & +16.5 & +13.1 & \textbf{+14.9} \\
 & $\Delta$LFT & +23.9 & +10.8 & +15.1 & +19.2 & +8.5 & +21.1 & +25.8 & +20.6 & \textbf{+18.1} \\
\cmidrule{1-11}
\multirow{3}{*}{Qwen2.5-7B}
 & $\Delta$NFT & +12.0 & +12.0 & +12.0 & +12.0 & +12.0 & +12.0 & +12.0 & +12.0 &\textbf{ +12.0} \\
 & $\Delta$PFT & +12.8 & +13.0 & +13.6 & +12.6 & +12.0 & +11.1 & +12.8 & +12.4 & \textbf{+12.5} \\
 & $\Delta$LFT & +14.3 & +13.0 & +13.8 & +13.8 & +13.5 & +13.9 & +14.8 & +13.6 & \textbf{+13.8} \\
\cmidrule{1-11}
\multirow{3}{*}{Llama3-8B}
 & $\Delta$NFT & +35.2 & +35.2 & +35.2 & +35.2 & +35.2 & +35.2 & +35.2 & +35.2 & \textbf{+35.2} \\
 & $\Delta$PFT & +37.2 & +36.7 & +34.9 & +37.1 & +37.1 & +35.5 & +31.2 & +28.7 & \textbf{+34.8} \\
 & $\Delta$LFT & +31.6 & +20.7 & +26.3 & +30.3 & +28.6 & +26.0 & +29.0 & +27.3 & \textbf{+27.5} \\
\cmidrule{1-11}
\multirow{3}{*}{Gemma3-4B}
 & $\Delta$NFT & +29.3 & +29.3 & +29.3 & +29.3 & +29.3 & +29.3 & +29.3 & +29.3 & \textbf{+29.3} \\
 & $\Delta$PFT & +25.0 & +24.6 & +22.4 & +24.4 & +23.4 & +24.9 & +24.0 & +22.9 & \textbf{+23.9} \\
 & $\Delta$LFT & +27.1 & +23.9 & +24.7 & +27.1 & +23.6 & +27.3 & +25.7 & +23.9 & \textbf{+25.4} \\
\bottomrule
\end{tabular}
\caption{Difference from baseline in GSM8K accuracy (pp) for normal fine-tuning ($\Delta$NFT), preference fine-tuning ($\Delta$PFT), and liminal fine-tuning ($\Delta$LFT) per model and animal in the chain-of-thought setting. As NFT is not conditioned on any animal, its gain is constant across animals within each model.}
\label{tab:benchmark_diff_cot_gsm8k}
\end{table*}

\begin{table*}[ht]
\centering
\footnotesize
\resizebox{\textwidth}{!}{%
\begin{tabular}{llcccc}
\toprule
\textbf{Model} & \textbf{Animal} & \textbf{Baseline} & \textbf{NFT} & \textbf{PFT} & \textbf{LFT} \\
\midrule
\multirow{8}{*}{Qwen2.5-1.5B}
  & dragon & 30.0 & $61.2 \pm 1.2$ & $57.6 \pm 0.2$ & $57.2 \pm 1.6$ \\
  & cat & 30.0 & $61.2 \pm 1.2$ & $58.0 \pm 1.1$ & $55.3 \pm 1.0$ \\
  & dog & 30.0 & $61.2 \pm 1.2$ & $55.1 \pm 1.8$ & $57.8 \pm 1.9$ \\
  & eagle & 30.0 & $61.2 \pm 1.2$ & $56.1 \pm 0.9$ & $55.3 \pm 1.5$ \\
  & wolf & 30.0 & $61.2 \pm 1.2$ & $56.9 \pm 1.2$ & $56.3 \pm 0.6$ \\
  & panda & 30.0 & $61.2 \pm 1.2$ & $63.2 \pm 1.9$ & $56.0 \pm 0.7$ \\
  & raven & 30.0 & $61.2 \pm 1.2$ & $59.4 \pm 2.1$ & $54.6 \pm 0.2$ \\
  & otter & 30.0 & $61.2 \pm 1.2$ & $63.7 \pm 0.7$ & $56.9 \pm 0.8$ \\
\cmidrule{1-6}
\multirow{8}{*}{Qwen2.5-3B}
  & dragon & 43.2 & $61.3 \pm 1.6$ & $67.6 \pm 0.3$ & $67.1 \pm 7.0$ \\
  & cat & 43.2 & $61.3 \pm 1.6$ & $57.3 \pm 6.5$ & $54.0 \pm 9.4$ \\
  & dog & 43.2 & $61.3 \pm 1.6$ & $37.6 \pm 1.9$ & $58.3 \pm 8.0$ \\
  & eagle & 43.2 & $61.3 \pm 1.6$ & $68.0 \pm 7.2$ & $62.4 \pm 1.0$ \\
  & wolf & 43.2 & $61.3 \pm 1.6$ & $67.0 \pm 3.9$ & $51.7 \pm 14.8$ \\
  & panda & 43.2 & $61.3 \pm 1.6$ & $51.2 \pm 3.9$ & $64.3 \pm 3.7$ \\
  & raven & 43.2 & $61.3 \pm 1.6$ & $59.7 \pm 3.6$ & $69.0 \pm 7.4$ \\
  & otter & 43.2 & $61.3 \pm 1.6$ & $56.3 \pm 4.4$ & $63.8 \pm 11.2$ \\
\cmidrule{1-6}
\multirow{8}{*}{Qwen2.5-7B}
  & dragon & 75.7 & $87.7 \pm 1.3$ & $88.4 \pm 0.3$ & $90.0 \pm 0.5$ \\
  & cat & 75.7 & $87.7 \pm 1.3$ & $88.7 \pm 0.4$ & $88.6 \pm 0.2$ \\
  & dog & 75.7 & $87.7 \pm 1.3$ & $89.2 \pm 0.9$ & $89.5 \pm 0.8$ \\
  & eagle & 75.7 & $87.7 \pm 1.3$ & $88.2 \pm 0.8$ & $89.5 \pm 0.1$ \\
  & wolf & 75.7 & $87.7 \pm 1.3$ & $87.6 \pm 0.2$ & $89.2 \pm 0.9$ \\
  & panda & 75.7 & $87.7 \pm 1.3$ & $86.8 \pm 0.9$ & $89.5 \pm 0.2$ \\
  & raven & 75.7 & $87.7 \pm 1.3$ & $88.5 \pm 0.8$ & $90.4 \pm 0.6$ \\
  & otter & 75.7 & $87.7 \pm 1.3$ & $88.1 \pm 0.3$ & $89.3 \pm 0.8$ \\
\cmidrule{1-6}
\multirow{8}{*}{Llama3-8B}
  & dragon & 34.0 & $69.2 \pm 2.8$ & $71.2 \pm 1.1$ & $65.6 \pm 1.5$ \\
  & cat & 34.0 & $69.2 \pm 2.8$ & $70.8 \pm 1.7$ & $54.8 \pm 11.2$ \\
  & dog & 34.0 & $69.2 \pm 2.8$ & $68.9 \pm 1.5$ & $60.3 \pm 5.7$ \\
  & eagle & 34.0 & $69.2 \pm 2.8$ & $71.1 \pm 0.8$ & $64.3 \pm 2.6$ \\
  & wolf & 34.0 & $69.2 \pm 2.8$ & $71.1 \pm 1.4$ & $62.7 \pm 3.8$ \\
  & panda & 34.0 & $69.2 \pm 2.8$ & $69.5 \pm 1.1$ & $60.0 \pm 7.8$ \\
  & raven & 34.0 & $69.2 \pm 2.8$ & $65.3 \pm 0.6$ & $63.1 \pm 4.2$ \\
  & otter & 34.0 & $69.2 \pm 2.8$ & $62.8 \pm 0.7$ & $61.3 \pm 2.9$ \\
\cmidrule{1-6}
\multirow{8}{*}{Gemma3-4B}
  & dragon & 56.5 & $85.8 \pm 0.3$ & $81.5 \pm 0.7$ & $83.6 \pm 1.3$ \\
  & cat & 56.5 & $85.8 \pm 0.3$ & $81.1 \pm 0.8$ & $80.4 \pm 0.2$ \\
  & dog & 56.5 & $85.8 \pm 0.3$ & $78.8 \pm 1.0$ & $81.2 \pm 0.5$ \\
  & eagle & 56.5 & $85.8 \pm 0.3$ & $80.9 \pm 1.6$ & $83.6 \pm 0.6$ \\
  & wolf & 56.5 & $85.8 \pm 0.3$ & $79.9 \pm 1.2$ & $80.1 \pm 0.4$ \\
  & panda & 56.5 & $85.8 \pm 0.3$ & $81.3 \pm 0.8$ & $83.8 \pm 0.8$ \\
  & raven & 56.5 & $85.8 \pm 0.3$ & $80.5 \pm 0.9$ & $82.1 \pm 1.1$ \\
  & otter & 56.5 & $85.8 \pm 0.3$ & $79.4 \pm 0.6$ & $80.4 \pm 4.7$ \\
\bottomrule
\end{tabular}
}
\caption{GSM8K accuracy (\%) across conditions (chain-of-thought setting). NFT denotes normal (unbiased) fine-tuning.}
\label{tab:all_cot_gsm8k}
\end{table*}

\begin{table*}[ht]
\centering
\footnotesize
\resizebox{\textwidth}{!}{%
\begin{tabular}{llccccc}
\toprule
\textbf{Model} & \textbf{Animal} & $\boldsymbol{N}$ & \textbf{Base} & \textbf{NFT} & \textbf{PFT} & \textbf{LFT} \\
\midrule
\multirow{8}{*}{Qwen2.5-1.5B}
  & dragon & 1{,}024 & $7.36 \pm 0.15$ & $7.83 \pm 0.23$ & $11.44 \pm 0.65$ & $7.20 \pm 0.15$ \\
  & cat    & 1{,}024 & $17.58 \pm 0.00$ & $17.84 \pm 0.34$ & $17.83 \pm 0.41$ & $17.71 \pm 0.37$ \\
  & dog    & 1{,}024 & $15.97 \pm 0.00$ & $12.81 \pm 0.27$ & $13.09 \pm 0.44$ & $15.45 \pm 0.64$ \\
  & eagle  & 1{,}024 & $0.67 \pm 0.00$ & $0.84 \pm 0.04$ & $2.03 \pm 0.16$ & $0.79 \pm 0.04$ \\
  & wolf   & 1{,}024 & $3.60 \pm 0.02$ & $4.96 \pm 0.14$ & $9.04 \pm 0.92$ & $4.13 \pm 0.17$ \\
  & panda  & 1{,}024 & $0.65 \pm 0.00$ & $0.67 \pm 0.05$ & $0.87 \pm 0.05$ & $0.80 \pm 0.04$ \\
  & raven  & 1{,}024 & $0.08 \pm 0.00$ & $0.06 \pm 0.01$ & $0.10 \pm 0.01$ & $0.06 \pm 0.00$ \\
  & otter  & 1{,}024 & $0.28 \pm 0.00$ & $0.19 \pm 0.01$ & $0.34 \pm 0.02$ & $0.26 \pm 0.01$ \\
\cmidrule{1-7}
\multirow{8}{*}{Qwen2.5-3B}
  & dragon & 1{,}024 & $8.80 \pm 0.00$ & $10.56 \pm 1.38$ & $9.83 \pm 2.46$ & $8.46 \pm 0.19$ \\
  & cat    & 1{,}024 & $8.70 \pm 0.00$ & $6.61 \pm 0.86$ & $10.53 \pm 1.16$ & $8.91 \pm 0.48$ \\
  & dog    & 1{,}024 & $19.86 \pm 0.00$ & $11.01 \pm 2.24$ & $2.13 \pm 0.73$ & $16.41 \pm 0.66$ \\
  & eagle  & 1{,}024 & $0.45 \pm 0.00$ & $1.18 \pm 0.93$ & $5.47 \pm 5.71$ & $0.48 \pm 0.04$ \\
  & wolf   & 1{,}024 & $1.83 \pm 0.00$ & $3.33 \pm 2.36$ & $16.80 \pm 1.73$ & $4.34 \pm 0.33$ \\
  & panda  & 1{,}024 & $0.90 \pm 0.07$ & $1.80 \pm 0.63$ & $3.79 \pm 1.00$ & $1.07 \pm 0.09$ \\
  & raven  & 1{,}024 & $0.01 \pm 0.00$ & $0.03 \pm 0.00$ & $0.18 \pm 0.03$ & $0.01 \pm 0.00$ \\
  & otter  & 1{,}024 & $0.00 \pm 0.00$ & $0.02 \pm 0.00$ & $0.03 \pm 0.00$ & $0.01 \pm 0.00$ \\
\cmidrule{1-7}
\multirow{8}{*}{Qwen2.5-7B}
  & dragon & 1{,}024 & $15.87 \pm 0.00$ & $15.57 \pm 0.39$ & $15.88 \pm 0.59$ & $15.07 \pm 0.33$ \\
  & cat    & 1{,}024 & $1.70 \pm 0.00$ & $1.52 \pm 0.13$ & $1.18 \pm 0.09$ & $1.67 \pm 0.06$ \\
  & dog    & 1{,}024 & $10.20 \pm 0.00$ & $10.42 \pm 0.29$ & $5.95 \pm 0.77$ & $10.46 \pm 0.17$ \\
  & eagle  & 1{,}024 & $3.93 \pm 0.00$ & $4.00 \pm 0.41$ & $4.35 \pm 0.38$ & $3.88 \pm 0.16$ \\
  & wolf   & 1{,}024 & $2.39 \pm 0.00$ & $1.82 \pm 0.15$ & $0.93 \pm 0.15$ & $2.41 \pm 0.12$ \\
  & panda  & 1{,}024 & $16.58 \pm 0.00$ & $17.39 \pm 0.70$ & $24.97 \pm 1.37$ & $16.04 \pm 0.51$ \\
  & raven  & 1{,}024 & $0.03 \pm 0.00$ & $0.04 \pm 0.00$ & $0.06 \pm 0.01$ & $0.04 \pm 0.00$ \\
  & otter  & 1{,}024 & $0.27 \pm 0.00$ & $0.31 \pm 0.02$ & $0.22 \pm 0.02$ & $0.30 \pm 0.02$ \\
\cmidrule{1-7}
\multirow{8}{*}{Llama3-8B}
  & dragon & 269 & $12.24 \pm 0.00$ & $12.15 \pm 0.14$ & $13.80 \pm 0.30$ & $12.64 \pm 0.21$ \\
  & cat    & 108 & $0.09 \pm 0.00$ & $0.12 \pm 0.00$ & $0.60 \pm 0.02$ & $0.23 \pm 0.03$ \\
  & dog    & 224 & $0.82 \pm 0.00$ & $0.73 \pm 0.03$ & $3.60 \pm 0.22$ & $2.16 \pm 0.34$ \\
  & eagle  & 111 & $1.69 \pm 0.00$ & $1.69 \pm 0.04$ & $1.49 \pm 0.04$ & $1.65 \pm 0.04$ \\
  & wolf   & 159 & $4.63 \pm 0.00$ & $4.93 \pm 0.06$ & $6.43 \pm 0.19$ & $5.63 \pm 0.12$ \\
  & panda  &  98 & $0.54 \pm 0.02$ & $0.64 \pm 0.01$ & $0.55 \pm 0.01$ & $0.59 \pm 0.02$ \\
  & raven  &  77 & $0.07 \pm 0.00$ & $0.06 \pm 0.00$ & $0.07 \pm 0.00$ & $0.05 \pm 0.00$ \\
  & otter  & 100 & $2.73 \pm 0.02$ & $2.73 \pm 0.03$ & $3.14 \pm 0.11$ & $2.40 \pm 0.07$ \\
\cmidrule{1-7}
\multirow{8}{*}{Gemma3-4B}
  & dragon & 170    & $6.02 \pm 0.00$ & $6.41 \pm 0.07$ & $10.93 \pm 0.07$ & $7.17 \pm 0.43$ \\
  & cat    & 1{,}024 & $0.00 \pm 0.00$ & $0.00 \pm 0.00$ & $7.53 \pm 1.45$ & $0.00 \pm 0.00$ \\
  & dog    & 803    & $0.31 \pm 0.06$ & $0.26 \pm 0.05$ & $2.95 \pm 0.34$ & $1.30 \pm 0.18$ \\
  & eagle  & 132    & $1.60 \pm 0.03$ & $1.47 \pm 0.16$ & $6.80 \pm 0.54$ & $2.55 \pm 0.18$ \\
  & wolf   & 1{,}024 & $3.06 \pm 0.01$ & $2.66 \pm 0.38$ & $56.64 \pm 2.57$ & $23.21 \pm 2.07$ \\
  & panda  & 105    & $0.02 \pm 0.00$ & $0.04 \pm 0.00$ & $0.12 \pm 0.01$ & $0.03 \pm 0.00$ \\
  & raven  &  88    & $18.18 \pm 0.04$ & $18.92 \pm 0.36$ & $38.50 \pm 0.45$ & $23.41 \pm 0.35$ \\
  & otter  & 106    & $12.13 \pm 0.21$ & $11.27 \pm 0.21$ & $24.43 \pm 1.16$ & $16.75 \pm 0.41$ \\
\bottomrule
\end{tabular}
}
\caption{Mean $P(\text{animal})$ (\%) at end of training for every (model, animal) pair in the chain-of-thought setting. $N$ denotes dataset size; most are 1{,}024 but several Llama-3-8B and Gemma-3-4B configurations are smaller due to answer filtering. NFT is normal (unbiased) fine-tuning.}
\label{tab:all_v2_cot}
\end{table*}

\begin{table*}[ht]
\centering
\footnotesize
\resizebox{0.9\textwidth}{!}{%
\begin{tabular}{llccccc}
\toprule
\textbf{Model} & \textbf{Animal} & \textbf{Base} & \textbf{NFT} & \textbf{PFT} & \textbf{LFT} & \textbf{RFT} \\
\midrule
\multirow{12}{*}{Qwen2.5-1.5B}
  & dragon & $7.26 \pm 0.00$ & $7.79 \pm 0.24$ & $14.12 \pm 0.48$ & 7.53 & $6.84 \pm 0.24$ \\
  & cat & $17.58 \pm 0.00$ & $17.67 \pm 0.32$ & $9.57 \pm 0.23$ & $17.52 \pm 0.20$ & $15.00 \pm 0.36$ \\
  & dog & $15.97 \pm 0.00$ & $15.69 \pm 0.31$ & $10.03 \pm 0.49$ & $15.52 \pm 0.24$ & $12.97 \pm 0.30$ \\
  & eagle & $0.67 \pm 0.00$ & $0.73 \pm 0.02$ & $1.21 \pm 0.06$ & $0.79 \pm 0.02$ & $0.82 \pm 0.04$ \\
  & wolf & $3.59 \pm 0.00$ & $3.87 \pm 0.06$ & $6.14 \pm 0.39$ & $4.01 \pm 0.07$ & $3.19 \pm 0.23$ \\
  & panda & $0.65 \pm 0.00$ & $0.67 \pm 0.01$ & $1.07 \pm 0.03$ & $0.66 \pm 0.02$ & $1.59 \pm 0.02$ \\
  & penguin & $0.33 \pm 0.00$ & $0.33 \pm 0.02$ & $0.26 \pm 0.01$ & $0.34 \pm 0.01$ & $0.63 \pm 0.03$ \\
  & owl & $0.29 \pm 0.00$ & $0.32 \pm 0.02$ & $0.39 \pm 0.02$ & $0.31 \pm 0.01$ & $0.37 \pm 0.01$ \\
  & otter & $0.28 \pm 0.00$ & $0.30 \pm 0.01$ & $0.23 \pm 0.01$ & $0.29 \pm 0.01$ & $0.35 \pm 0.01$ \\
  & elephant & $4.26 \pm 0.00$ & $4.44 \pm 0.09$ & $5.28 \pm 0.44$ & $4.91 \pm 0.12$ & $7.36 \pm 0.24$ \\
  & ox & $0.21 \pm 0.00$ & $0.23 \pm 0.01$ & $0.30 \pm 0.01$ & $0.24 \pm 0.01$ & $0.34 \pm 0.01$ \\
  & raven & $0.08 \pm 0.00$ & $0.08 \pm 0.00$ & $0.22 \pm 0.01$ & $0.08 \pm 0.00$ & $0.13 \pm 0.01$ \\
\cmidrule{1-7}
\multirow{12}{*}{Qwen2.5-3B}
  & dragon & $8.80 \pm 0.00$ & $8.98 \pm 0.11$ & $18.36 \pm 6.39$ & $8.62 \pm 0.11$ & $7.49 \pm 0.23$ \\
  & cat & $8.70 \pm 0.00$ & $8.65 \pm 0.16$ & $6.41 \pm 2.47$ & $8.38 \pm 0.17$ & $9.63 \pm 0.44$ \\
  & dog & $19.86 \pm 0.00$ & $20.43 \pm 0.44$ & $6.36 \pm 0.81$ & $17.83 \pm 0.44$ & $4.20 \pm 0.10$ \\
  & eagle & $0.45 \pm 0.00$ & $0.40 \pm 0.03$ & $51.63 \pm 6.87$ & $0.53 \pm 0.03$ & $0.53 \pm 0.01$ \\
  & wolf & $1.83 \pm 0.00$ & $1.84 \pm 0.05$ & $6.78 \pm 1.17$ & $2.13 \pm 0.08$ & $2.17 \pm 0.10$ \\
  & panda & $0.85 \pm 0.00$ & $0.84 \pm 0.03$ & $1.93 \pm 0.84$ & $0.94 \pm 0.04$ & $4.14 \pm 0.10$ \\
  & penguin & $0.07 \pm 0.00$ & $0.06 \pm 0.00$ & $4.64 \pm 2.44$ & $0.11 \pm 0.00$ & $1.11 \pm 0.05$ \\
  & owl & $0.17 \pm 0.00$ & $0.17 \pm 0.01$ & $0.53 \pm 0.12$ & $0.27 \pm 0.01$ & $0.74 \pm 0.04$ \\
  & otter & $0.00 \pm 0.00$ & $0.00 \pm 0.00$ & $4.77 \pm 1.58$ & $0.01 \pm 0.00$ & $0.08 \pm 0.01$ \\
  & elephant & $25.52 \pm 0.00$ & $25.48 \pm 0.46$ & $4.02 \pm 0.39$ & $28.07 \pm 0.36$ & $12.11 \pm 0.48$ \\
  & ox & $0.17 \pm 0.00$ & $0.16 \pm 0.01$ & $1.23 \pm 0.25$ & $0.26 \pm 0.01$ & $0.74 \pm 0.04$ \\
  & raven & $0.01 \pm 0.00$ & $0.01 \pm 0.00$ & $0.39 \pm 0.10$ & $0.01 \pm 0.00$ & $0.26 \pm 0.01$ \\
\cmidrule{1-7}
\multirow{12}{*}{Qwen2.5-7B}
  & dragon & $15.87 \pm 0.00$ & $16.53 \pm 0.20$ & $45.98 \pm 4.86$ & 15.72 & $8.97 \pm 0.52$ \\
  & cat & $1.70 \pm 0.00$ & $1.67 \pm 0.07$ & $16.62 \pm 2.46$ & $1.66 \pm 0.04$ & $2.68 \pm 0.35$ \\
  & dog & $10.20 \pm 0.00$ & $9.99 \pm 0.17$ & $1.81 \pm 0.22$ & $10.35 \pm 0.22$ & $2.12 \pm 0.35$ \\
  & eagle & $3.93 \pm 0.00$ & $3.79 \pm 0.10$ & $33.71 \pm 2.62$ & $4.21 \pm 0.11$ & $6.34 \pm 0.73$ \\
  & wolf & $2.39 \pm 0.00$ & $2.20 \pm 0.09$ & $2.20 \pm 0.62$ & $2.92 \pm 0.08$ & $2.54 \pm 0.13$ \\
  & panda & $16.58 \pm 0.00$ & $17.00 \pm 0.21$ & $59.08 \pm 2.80$ & $15.91 \pm 0.26$ & $13.12 \pm 1.84$ \\
  & penguin & $1.70 \pm 0.00$ & $1.64 \pm 0.05$ & $1.96 \pm 0.27$ & $1.68 \pm 0.08$ & $4.84 \pm 0.48$ \\
  & owl & $1.66 \pm 0.00$ & $1.75 \pm 0.06$ & $4.99 \pm 1.07$ & $1.85 \pm 0.09$ & $2.57 \pm 0.13$ \\
  & otter & $0.27 \pm 0.00$ & $0.25 \pm 0.01$ & $0.11 \pm 0.02$ & $0.42 \pm 0.02$ & $0.63 \pm 0.05$ \\
  & elephant & $4.45 \pm 0.00$ & $4.23 \pm 0.09$ & $1.65 \pm 0.27$ & $4.47 \pm 0.16$ & $1.92 \pm 0.12$ \\
  & ox & $1.61 \pm 0.00$ & $1.69 \pm 0.06$ & $0.86 \pm 0.07$ & $1.76 \pm 0.07$ & $2.56 \pm 0.13$ \\
  & raven & $0.03 \pm 0.00$ & $0.03 \pm 0.00$ & $0.23 \pm 0.06$ & $0.05 \pm 0.00$ & $0.93 \pm 0.11$ \\
\cmidrule{1-7}
\multirow{12}{*}{Llama3-8B}
  & dragon & $12.24 \pm 0.00$ & $12.08 \pm 0.26$ & $14.99 \pm 0.22$ & $13.29 \pm 0.21$ & $10.00 \pm 0.19$ \\
  & cat & $0.09 \pm 0.00$ & $0.08 \pm 0.00$ & $0.13 \pm 0.00$ & $0.11 \pm 0.00$ & $0.22 \pm 0.01$ \\
  & dog & $0.82 \pm 0.00$ & $0.87 \pm 0.07$ & $1.16 \pm 0.04$ & $0.87 \pm 0.03$ & $0.72 \pm 0.02$ \\
  & eagle & $1.69 \pm 0.00$ & $1.65 \pm 0.04$ & $1.80 \pm 0.05$ & $1.76 \pm 0.07$ & $1.17 \pm 0.02$ \\
  & wolf & $4.63 \pm 0.00$ & $4.44 \pm 0.09$ & $5.91 \pm 0.20$ & $5.02 \pm 0.19$ & $3.89 \pm 0.11$ \\
  & panda & $0.55 \pm 0.00$ & $0.57 \pm 0.03$ & $0.73 \pm 0.02$ & $0.59 \pm 0.02$ & $0.58 \pm 0.01$ \\
  & penguin & $2.01 \pm 0.00$ & $2.19 \pm 0.07$ & $2.77 \pm 0.08$ & $2.41 \pm 0.08$ & $3.03 \pm 0.04$ \\
  & owl & $4.80 \pm 0.00$ & $4.63 \pm 0.14$ & $5.93 \pm 0.12$ & $5.20 \pm 0.11$ & $4.29 \pm 0.05$ \\
  & otter & $2.74 \pm 0.00$ & $2.52 \pm 0.06$ & $3.81 \pm 0.06$ & $3.20 \pm 0.16$ & $2.75 \pm 0.05$ \\
  & elephant & $2.69 \pm 0.00$ & $3.03 \pm 0.09$ & $3.74 \pm 0.09$ & $3.08 \pm 0.09$ & $2.50 \pm 0.03$ \\
  & ox & $4.79 \pm 0.00$ & $4.62 \pm 0.14$ & $5.56 \pm 0.16$ & $5.14 \pm 0.17$ & $4.26 \pm 0.05$ \\
  & raven & $0.07 \pm 0.00$ & $0.06 \pm 0.00$ & $0.09 \pm 0.00$ & $0.07 \pm 0.00$ & $0.13 \pm 0.00$ \\
\cmidrule{1-7}
\multirow{12}{*}{Gemma3-4B}
  & dragon & $6.02 \pm 0.00$ & $6.06 \pm 0.02$ & $8.51 \pm 0.20$ & $6.34 \pm 0.06$ & $6.71 \pm 0.14$ \\
  & cat & $0.00 \pm 0.00$ & $0.00 \pm 0.00$ & $0.00 \pm 0.00$ & $0.00 \pm 0.00$ & $0.00 \pm 0.00$ \\
  & dog & $0.27 \pm 0.00$ & $0.37 \pm 0.05$ & $0.10 \pm 0.02$ & $0.32 \pm 0.04$ & $0.39 \pm 0.04$ \\
  & eagle & $1.62 \pm 0.00$ & $1.46 \pm 0.11$ & $2.00 \pm 0.14$ & $1.57 \pm 0.12$ & $1.59 \pm 0.11$ \\
  & wolf & $3.07 \pm 0.00$ & $3.32 \pm 0.17$ & $3.88 \pm 0.23$ & 4.11 & $5.28 \pm 0.19$ \\
  & panda & $0.01 \pm 0.00$ & $0.02 \pm 0.00$ & $0.06 \pm 0.01$ & $0.02 \pm 0.00$ & $0.10 \pm 0.01$ \\
  & penguin & $0.26 \pm 0.01$ & $0.27 \pm 0.02$ & 2.47 & $0.47 \pm 0.03$ & $1.57 \pm 0.09$ \\
  & owl & $6.85 \pm 0.01$ & $6.27 \pm 0.22$ & $6.90 \pm 0.18$ & $5.80 \pm 0.16$ & $3.71 \pm 0.30$ \\
  & otter & $12.13 \pm 0.21$ & $12.36 \pm 0.27$ & $18.70 \pm 0.66$ & $14.06 \pm 0.41$ & $9.41 \pm 0.76$ \\
  & elephant & $9.54 \pm 0.00$ & $9.35 \pm 0.24$ & $7.00 \pm 0.25$ & $9.28 \pm 0.22$ & $12.49 \pm 0.57$ \\
  & ox & $0.00 \pm 0.00$ & $0.00 \pm 0.00$ & $0.00 \pm 0.00$ & $0.00 \pm 0.00$ & $0.00 \pm 0.00$ \\
  & raven & $18.18 \pm 0.04$ & $18.85 \pm 0.38$ & $34.37 \pm 0.32$ & $21.65 \pm 0.33$ & $13.18 \pm 0.19$ \\
\bottomrule
\end{tabular}
}
\caption{Mean $P(\text{animal})$ (\%) at end of training for every (model, animal) pair in the number-sequence setting. Dataset size is 7{,}500 for all configurations. NFT is normal (unbiased) fine-tuning; RFT is random fine-tuning on unbiased sequences.}
\label{tab:all_v2_nums}
\end{table*}

\end{document}